\documentclass[a4paper,fleqn]{caswenjia}

\usepackage[authoryear]{natbib}

\usepackage{amsmath}
\usepackage{amsfonts}
\usepackage{amssymb}
\usepackage{bbm}
\usepackage{hyperref}

\usepackage{caption}

\DeclareMathOperator*{\argmax}{arg\,max}

\newcommand{\ie}{\textit{i.e.}}
\newcommand{\eg}{\textit{e.g.}}

\newcommand{\revise}[1]{\textcolor{black}{#1}}
\newcommand{\tworevise}[1]{\textcolor{black}{#1}}

\usepackage{amsmath}
\newcommand*{\Scale}[2][4]{\scalebox{#1}{$#2$}}%

\def\tsc#1{\csdef{#1}{\textsc{\lowercase{#1}}\xspace}}
\tsc{WGM}
\tsc{QE}
\tsc{EP}
\tsc{PMS}
\tsc{BEC}
\tsc{DE}

\begin{document}
\begin{sloppypar}

\let\WriteBookmarks\relax
\def\floatpagepagefraction{1}
\def\textpagefraction{.001}
\shorttitle{Deep Semantic-Visual Alignment for Zero-Shot Remote Sensing Image Scene Classification}
\shortauthors{Wenjia Xu et~al.}

\title [mode = title]{Deep Semantic-Visual Alignment for Zero-Shot Remote Sensing Image Scene Classification}                      




\author[1,3,4]{Wenjia Xu}[orcid=0000-0002-1425-4162]
\ead{xuwenjia@bupt.edu.cn}

\credit{Conceptualization of this study, Methodology, Software}

\address[1]{State Key Laboratory of Networking and Switching Technology, Beijing University of Posts and Telecommunications, Beijing 100876, China}

\author[2,3,4]{Jiuniu Wang}
\cormark[1]

\author[3,4]{Zhiwei Wei}

\author[1]{Mugen Peng}

\author[3,4]{Yirong Wu}

\credit{Data curation, Writing - Original draft preparation}
\address[2]{City University of Hong Kong, Hong Kong 999077, China}
\address[3]{University of Chinese Academy of Sciences, Beijing 100864, China}
\address[4]{Aerospace Information Research Institute, CAS, Beijing 100094, China}

\cortext[cor1]{Corresponding author}


\begin{abstract}
Deep neural networks have achieved promising progress in remote sensing (RS) image classification, for which the training process requires abundant samples for each class. However, it is time-consuming and unrealistic to annotate labels for each RS category, given the fact that the RS target database is increasing dynamically.  Zero-shot learning~(ZSL) allows for identifying novel classes that are not seen during training, which provides a promising solution for the aforementioned problem. However, previous ZSL models mainly depend on manually-labeled attributes or word embeddings extracted from language models to transfer knowledge from seen classes to novel classes. Those class embeddings may not be visually detectable and the annotation process is time-consuming and labor-intensive. Besides, pioneer ZSL models use convolutional neural networks pre-trained on ImageNet, which focus on the main objects appearing in each image, neglecting the background context that also matters in RS scene classification. \tworevise{To address the above problems, we propose to collect visually detectable attributes automatically. We predict attributes for each class by depicting the semantic-visual similarity between attributes and images. In this way, the attribute annotation process is accomplished by machine instead of human as in other methods.} Moreover, we propose a Deep Semantic-Visual Alignment~(DSVA) that take advantage of the self-attention mechanism in the transformer to associate local image regions together, integrating the background context information for prediction. The DSVA model further utilizes the attribute attention maps to focus on the informative image regions that are essential for knowledge transfer in ZSL, and maps the visual images into attribute space to perform ZSL classification. With extensive experiments, we show that our model outperforms other state-of-the-art models by a large margin on a challenging large-scale RS scene classification benchmark.  Moreover, we qualitatively verify that the attributes annotated by our network are both class discriminative and semantic related, which benefits the zero-shot knowledge transfer. The code is available at 
\url{https://github.com/wenjiaXu/RS_Scene_ZSL}.
\end{abstract}




\maketitle
\section{Introduction}

With the rapid advances of sensors and remote sensing (RS) technology~\citep{toth2016remote}, RS image scene classification~\citep{cheng2017remote,gu2019survey} draws increasing attention as it is playing an essential role in urban construction~\citep{chen2021identification}, environmental monitoring~\citep{li2020review}, etc. 
While deep neural networks (DNNs) have led to impressive successes in image scene classification~\citep{cheng2020remote,wang2020looking}, learning the hyperplane for discriminating various categories requires abundant training samples~\citep{39_Imagenet,cheng2020remote}. However, it is unrealistic to collect sufficient RS scene images for all circumstances~\citep{xue2017sparse,alajaji2020few,li2020rs}. For instance, the earth observation system can collect a huge size of data (up to 100TB) every day, while it is notably time consuming to annotate all category labels at once. Moreover, as the RS target database is increasing dynamically, there is an urgent demand for recognizing new RS scenes that never appear in the training stage.

\begin{figure}[tb]
    \centering
    \includegraphics[width=.95\linewidth]{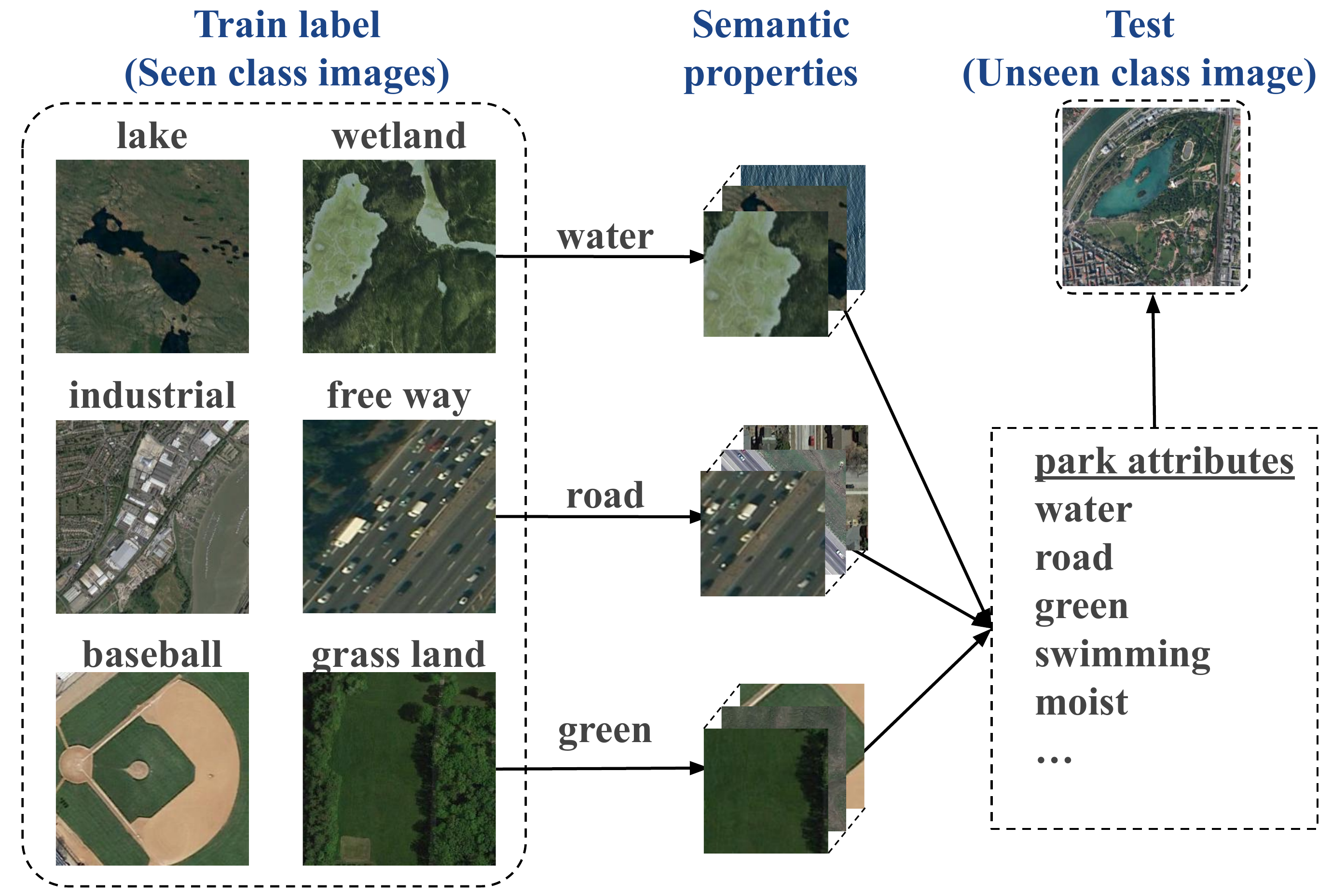}
    \caption{Illustration of zero-shot learning. By learning the semantic properties, \eg, ``water'', ``road'', from seen classes such as lake and wetland, the model would be able to recognize the unseen class ``park'' composing of those semantic properties.}
    \label{fig:teaser}
\end{figure}

Zero-shot learning~(ZSL), which aims to identify unseen classes without training samples~\citep{xian2018zero}, provides a promising solution for the aforementioned problem, and are widely explored in image classification~\citep{xu2022vgse}, object detection~\citep{zhu2019zero}, etc. By leveraging semantic knowledge and visual information for each seen category, ZSL models can generalize the learned knowledge to unseen classes simultaneously. As shown in Figure~\ref{fig:teaser}, by learning the visual property of semantic knowledge demonstrating each seen class (\eg, what is ``water'', ``road'', and ``green''), the model would be able to recognize unseen category \textit{park} composing of that semantic knowledge. The two key factors of ZSL are the class embeddings and the ZSL model~\citep{xu2020attribute}. The class embeddings aggregated for every class live in a semantic vector space that can be used to associate different classes even when visual examples of these classes are not available. Meanwhile, the ZSL model learns how to transfer knowledge from seen classes to unseen classes. 



The prior attempts in improving class embeddings are two folds, manually-labeled attributes and semantic embeddings from pre-trained language models. 
Attributes are the characteristic properties of objects~\citep{25_SUNdataset,26_wah2011caltech,farhadi2009describing}, which are both human interpretable and discriminative among various categories. Therefore, attribute embeddings describing the characteristics of each class have become the most widely used and powerful class embeddings for ZSL~\citep{xian2018zero,xu2022attribute,ALE}. Obtaining manually-labeled attributes is often a two-step process~\citep{25_SUNdataset}, which is time-consuming and labor-intensive. First, domain experts carefully design an attribute vocabulary containing various attributes, \eg, color, shape, etc. 
Second, human annotators would check each category and annotate the presence or absence of an attribute in an image or a class, which is called the labeling process. Although there are some attempts to collect attributes for remote sensing scenarios, \eg, for aircraft recognition~\citep{xu2020model} and scene classification~\citep{li2021robust}, those attributes are either incomplete in visual space~(with less than 60 attributes per class), or require enormous human efforts. Some pioneers tackle this problem by replacing attributes with semantic class embeddings extracted from pre-trained language models, \eg, word2vec~\citep{w2v}, glove~\citep{glove}, and BERT~\citep{devlin2018bert}, or from knowledge graphs constructed for RS scenes~\citep{li2021robust}. However, each dimension of these embeddings does not contain concrete semantic properties, and cannot be visually detectable, thus resulting in inferior ZSL performance. Therefore, to promote the development of ZSL in the remote sensing era, there is an urgent need to collect rich attributes depicting the visual properties of RS categories, and build a manual-free attribute labeling system that can collect visually detectable attributes. 

\revise{Since RS images are usually collected from satellites or aircraft, while ordinary optical images are collected from the ground, the image distribution between the two differs from each other in the following two aspects. The generalization of ordinary ZSL models to the RS scene classification task is limited due to the following characteristics of the RS scene. First, dislike ordinary optical images that usually focus on the local objects~\citep{39_Imagenet,xu2020model}, both the global context and the interaction information between local objects are important. As shown in Figure~\ref{fig:teaser}, the road, buildings, and the background grasses all matter in recognizing the scene as the \textit{park} instead of \textit{industrial} or \textit{grass lands}. Second, RS images have high inter-class variance and intra-class similarity~\citep{zhao2021hyperspectral}, thus requiring the ZSL models to concentrate on the intra-class discriminative features and the inter-class shared features~\citep{cheng2017remote}.} However, most of the remote sensing ZSL models~\citep{li2017zero,wang2021distance,li2021learning,li2021robust} utilize convolutional neural network~(CNN)~\citep{he2016deep} pre-trained on large-scale ordinary image dataset, \eg, ImageNet~\citep{39_Imagenet}, ignoring the domain gap between the ordinary and RS images. Furthermore, CNN networks inherently have smaller receptive fields by design, which focus on local object features and cannot fully correlate global and local information for large-scale RS scenes.



To tackle the key problems analyzed above, we propose two networks to improve the quality of RS attribute embeddings and the ZSL model, which integrates semantic and visual information to boost the ZSL performance. \tworevise{First, we propose an automatic attribute annotation process, where we use the CLIP model~\citep{radford2021learning} to predict remote sensing multi-modal attributes~(denoted as RSMM-Attributes) for remote sensing scenes. We first construct an attribute vocabulary containing attributes with rich semantic and visual information for RS scene classes.} To ensure the attribute embeddings for each class are visually detectable, the attribute labeling process is finished by calculating the semantic-visual similarity between the attribute and the example images. The CLIP model is pre-trained with a large-scale RS dataset containing images and the corresponding text descriptions, thus being able to associate the semantic text and visual RS images in one common space. In this way, the labor-intensive labeling process is replaced with an automatic model, significantly decreasing the time consumption.

To tackle the problems of remote sensing ZSL tasks, we develop a Deep Semantic-Visual Alignment~(DSVA) model that uses the RSMM-Attributes to transfer knowledge between seen and unseen classes, and perform ZSL considering both the local details and global contexts of images. We adopt a \revise{vision transformer}~\citep{dosovitskiy2020image}, which use long-range interaction between local image region to extract image representations. \revise{The self-attention mechanism~\citep{vaswani2017attention} in transformer helps to correlate the global context and local information together, which helps to integrate global information for ZSL. This is critical for remote sensing scene recognition as the global interaction of different image areas contributes to scene prediction.} \revise{To associate semantic attributes with visual features, we then learn attribute prototypes that encode the visual properties for each attribute.} Meanwhile, we generate the attribute attention maps by calculating the similarity between prototypes and the visual image features and accurately map the image into attribute space by calculating the image-attribute similarity. We further propose an attention concentration module to focus on the informative attribute regions. \revise{Since attributes represent intra-class discriminative features and inter-class shared features, this module will help the network to take advantage of discriminative image regions and benefit the knowledge transfer in ZSL.} 

In summary, the contributions of this work are three folds.
\begin{itemize}
    \item \tworevise{We propose to automatically predict visual attributes for remote sensing scenes, alleviating the manual effort needed for labeling attributes. Our RSMM-Attributes are visually detectable and can facilitate the knowledge transfer between seen and unseen classes.}
    
    \item We elaborate on the difference between ordinary images and remote sensing images, and develop a Deep Semantic-Visual Alignment~(DSVA) model that uses RSMM-Attributes to tackle the RS zero-shot learning problem. Our model adopts a \revise{vision transformer} with self-attention mechanism to enlarge the receptive field and learn both the local details and global contexts for RS scenes. Furthermore, an attention concentration module is proposed to focus on the informative attribute regions.
    
    \item Extensive quantitative experiments demonstrate that our model achieves state-of-the-art performance on a large-scale RS scene benchmark, outperforming the other pioneers by up to 30\%. Qualitative analysis indicates that the learned RSMM-Attributes are both class discriminative and can reflect the visual and semantic relatedness.
\end{itemize}

The remainder of the paper is organized as follows. In Section~\ref{sec:Related}, we present related works, including ZSL models and class embeddings. In Section~\ref{sec:RSMM}, we introduce the process for collecting visually detectable and automatic-labeled attributes for RS scenes. In Section~\ref{sec:DSVA} we introduce our Deep Semantic-Visual Alignment~(DSVA) model that performs ZSL considering the global context of RS scenes. The experimental setting and quantitative results are presented in Section \ref{sec:Exp}, demonstrating the superior performance of our work. In Section~\ref{sec:vis}, we present the visualization of the learnt RSMM-Attributes. Finally, we conclude the paper in Section~\ref{sec:Conclu}.









\section{Related Works}
\label{sec:Related}
In this section, we review the related literature concerning the two key factors in ZSL, \ie, the zero-shot learning models and the class embeddings.

\subsection{Zero-shot learning in remote sensing scene classification}
In ZSL, the goal is to train a model with abundant training samples from seen classes, then recognize unseen classes that are not observed during training. This is accomplished by using class embeddings, also known as side information~\citep{xian2018zero}, that describe the semantic properties among seen and unseen classes. One of the earliest attempts in ZSL can be traced to \citet{lampert2009learning} who perform direct attribute prediction for unseen classes. After that, approaches in this direction can be divided into two groups. The first group originates from attribute latent embeddings~(ALE)~\citep{ALE}, where the model maps the image representations to the space of class embeddings, and learns a compatibility function between them. Some following notable works aim at learning better embedding spaces~\citep{zhang2016zero}, improving the compatibility functions~\citep{liu2021goal}, or enhancing the image encoders~\citep{xu2022attribute}. An alternative line of work argues that the zero-shot problem can be complemented by generating fake samples for unseen classes~\citep{xian2018feature}. This is achieved by synthesizing image representations for unseen classes with generative models, \eg,  generative adversarial networks~(GAN)~\citep{goodfellow2014generative} and variational auto encoder~(VAE)~\citep{kingma2013auto}, conditioned on the class embeddings~\citep{ABP,schonfeld2019generalized}. 

The first adaptation in the RS field can be traced to \citet{li2017zero} who utilize an unsupervised domain adaptation model to predict the unseen class labels, and develop a label propagation algorithm to leverage the visual similarity among images from the same scene class. \citet{sumbul2017fine} further utilize a bilinear function that models the compatibility between the visual images and the class embeddings. Instead of using image features extracted from a fixed CNN network, LPDCMENs~\citep{li2021learning} train the network end-to-end and preserve the locality in RS scene images by emphasizing the pairwise similarity in one class. To maintain the cross-modal alignment between visual and semantic space, \citet{li2021robust} propose a novel generative model DAN to synthesize image features and match semantic features in a latent space, and \citet{wang2021distance} propose an autoencoder model (DSAE) constrained by Euclidean distance between samples. 

Despite the rapid development of ZSL models in the field of computer vision, the unique characteristics of remote sensing images limit the generalization of the above models in remote sensing ZSL tasks. First, since RS images usually have high inter-class variance and high intra-class similarity, the ZSL model designed for ordinary optical images cannot discriminate between similar categories.
Second, while above approaches improve the ZSL performance gradually, they all utilize CNN as the backbone to extract image features, and ignore the intrinsic difference between RS images and ordinary optical images, \ie, the global information. Dislike ordinary images that usually contain forehead regions that should draw much attention and background information that could be discarded, there is not an obvious distinction between foreground and background in RS images and every pixel matters when performing scene classification task. To this end, we adopt a \revise{vision transformer} with self-attention mechanism to enlarge the receptive field and learn both the local details and the global contexts of RS scenes. Besides, our DSVA network is encoded with an attention concentration module focusing on the attribute-related informative image regions, which is helpful for discriminating between different classes.

\subsection{Class embeddings for zero-shot learning}
%
Class embeddings are crucial in relating different categories with shared characteristics in the semantic space, and can transfer knowledge from seen classes to unseen classes in ZSL. There are three types of commonly used class embeddings. The human annotated attributes~\citep{25_SUNdataset,farhadi2009describing}, describing the visual and semantic properties of objects, are the most popular class embeddings in zero-shot learning. Though attributes show powerful capability in discriminating and associating different classes, they are limited as the annotation process is time-consuming and labor-intensive~\citep{xu2022vgse}. An alternative to the manual attributes is to extract class embeddings using pre-trained language models such as BERT~\citep{devlin2018bert}, and word2vec~\citep{w2v}. Other works utilize knowledge graphs~\citep{nayak2020zero} and online encyclopedia~\citep{al2017automatic} to extract class embeddings to encode more knowledge for each category. 

Despite their importance, class embeddings are relatively under-explored in zero-shot learning RS scene classification. Previous works mainly use class embeddings extracted from pre-trained language models, or attributes manually-labeled by experts. \citet{li2017zero} directly leverage pre-trained word2vec model to map the name of RS scene category to the semantic space, and \citet{li2022generative} further investigate more language models such as fasttext~\citep{bojanowski2017enriching} and BERT~\citep{devlin2018bert}. Although word vector can work as class embeddings, they are typically not comprehensive in capturing the visual properties of categories, and the following works improve this situation with the help of domain experts. \citet{sumbul2017fine} collect 25 attributes determining visually distinctive characters of each category, such as their parts, texture, and shape. \citet{li2021learning} ask multiple domain experts who are familiar with the RS field to observe 10 images from each category and summarize them with one sentence. The class embeddings are then extracted from a pre-trained BERT model. \citet{li2021robust} collect 59 attributes for 700 remote sensing scenes considering the color, shape, and objects. They further construct an RS scene knowledge graph with the help of 10 domain experts. However, the attributes and knowledge graphs cannot fully describe the rich visual properties of each category, and the time-consuming labeling process limits the generalization to new classes in practical. To this end, we propose to alleviate the manual burden by replacing the manual labeling process with an remote sensing multi-modal network. Our network depicts a few images and annotates the attribute value for each category automatically by measuring the similarity between the attributes and the images, which ensures that the annotated attributes can be visually detectable and describes the visual properties for all categories.

\begin{figure}[tb]
    \centering
    \includegraphics[width=.95\linewidth]{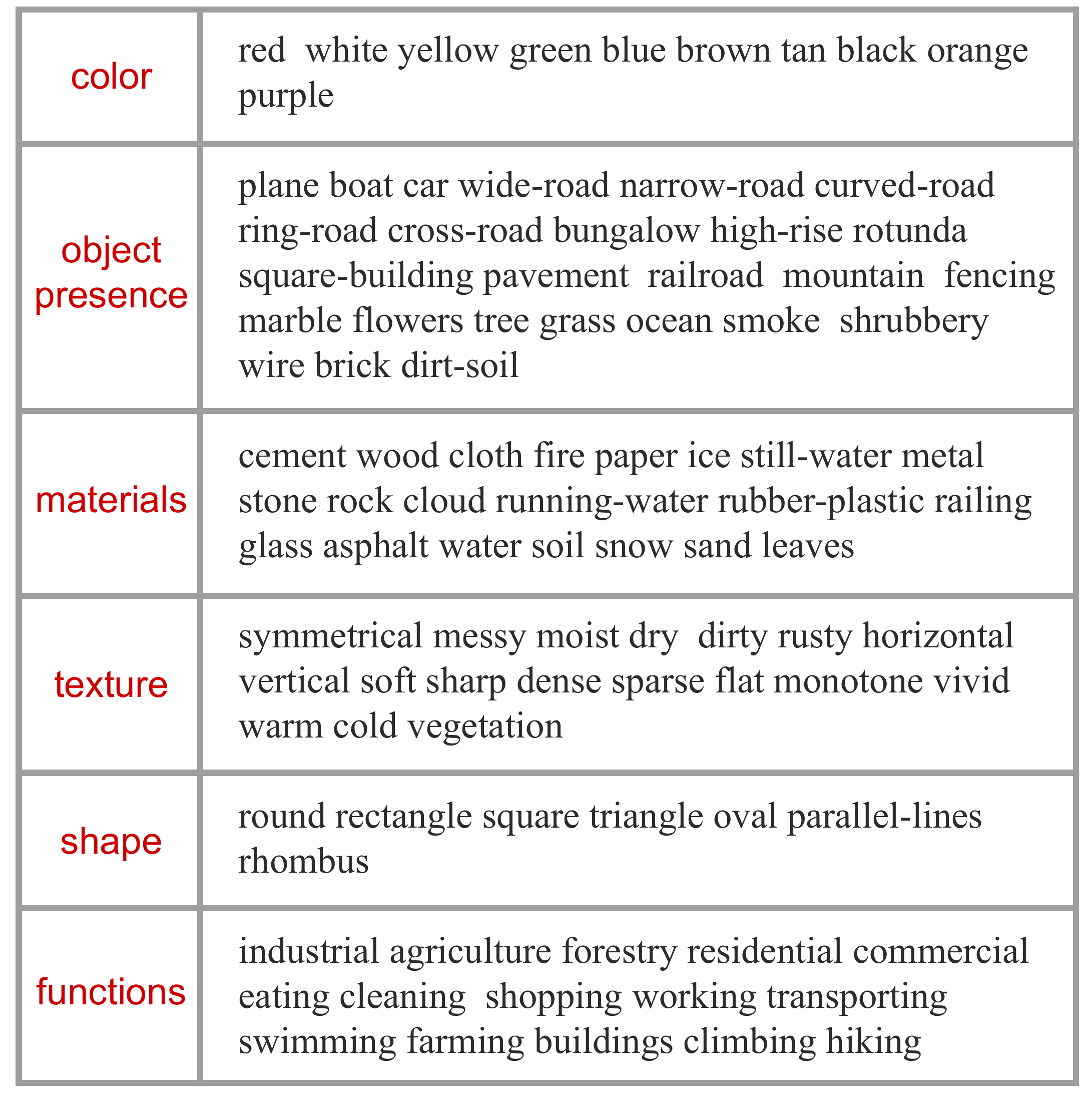}
    \caption{Attribute vocabulary. We split 98 attributes into 6 groups, \ie, ``color'', ``object presence'', ``material'', ``texture'', ``shape'', ``functions''.}
    \label{fig:attribute}
\end{figure}

\section{\tworevise{Automatic attribute annotation with remote sensing multi-modal similarity}}
\label{sec:RSMM}


In this section, we are interested in the automatic attribute annotation process considering the rich visual information in remote sensing scenes while reducing human labor in attribute labeling. We would first construct an attribute vocabulary $\mathcal{A}$ covering the semantic and visual properties of all remote sensing scenes. \tworevise{Then we propose to adapt the CLIP~\citep{radford2021learning} model which links the semantic-visual space together to annotate the Remote Sensing Multi-Modal Attribute~(RSMM-Attribute) for each remote sensing scene category.} 

\tworevise{Given a remote sensing scene category $y$ and one attribute $a \in \mathcal{A}$, the purpose is to annotate the attribute value for this category as $r_a(y) \in \mathbb{R}$, indicating the possibility that the attribute $a$ appearing in class $y$. With the CLIP model, we can measure the strength of association $r_a(y)$ between the attribute $a$ and category $y$. In the end, we will get a class embedding containing all attribute values for each category as $r_{\mathcal{A}}(y) \in \mathbb{R}^{N_a}$, where $N_a$ is the size of attribute vocabulary. In this section, we would first introduce the procedure of attribute vocabulary construction and automatic attribute annotation, then introduce how the CLIP model is trained and fine-tuned.}

\begin{figure*}
    \centering
    \includegraphics[width=\linewidth]{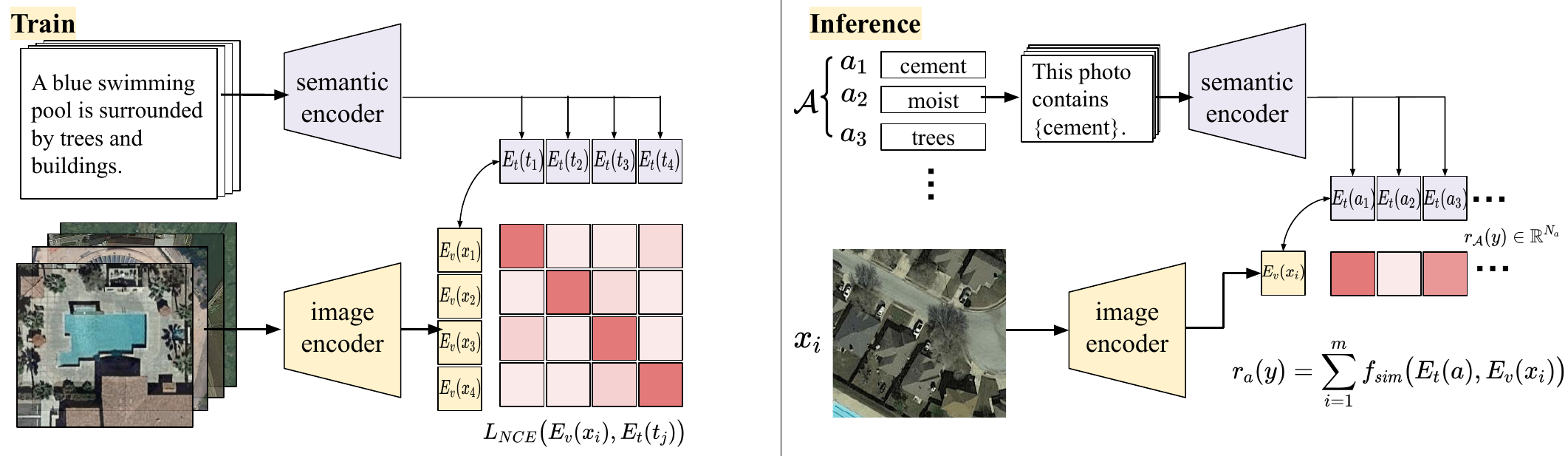}
    \caption{\tworevise{The attribute annotation process. Left: We show the fine-tune process with a batch of 4 images and the corresponding text descriptions. Right: The inference process of CLIP network where we measure the similarity between the probe image $x_i$ and attributes.}}
    \label{fig:RSMM}
\end{figure*}

\subsection{Attribute vocabulary construction}

To discover attributes with rich semantic and visual information, we consider the following six types of attributes as shown in Figure~\ref{fig:attribute}. (1) The ``color'' group includes the color appearing in the scene~(\eg, green, brown). (2) The ``object presence'' group lists the objects that would show up~(\eg, tree, soil). (3) The ``materials'' group describes the material constructing the scene~(\eg, cement, metal). (4) The ``texture'' group describes the texture and pattern of each category~(\eg, symmetrical, flat). (5) The ``shape'' group indicates the main shape shown up in each scene~(\eg, round, rectangle). (6) The \revise{``functions''} group indicates the social-economic function of each category~(\eg, industrial, agriculture). Then we go through all the properties and objects related to the remote sensing scene to fill each group, and finally get an attribute vocabulary $\mathcal{A}$ with ${N_a}$ attributes.
Note that building the attribute vocabulary is relatively easy, with only one annotator working for three hours. Besides, the vocabulary is not limited to the known scene categories, as the attributes can be shared between classes, therefore can be generalized to novel RS scene categories and describe their discriminate properties.



\subsection {Automatic attribute annotation with CLIP model}
\tworevise{As shown in Figure~\ref{fig:RSMM}~(right), the CLIP model we adopt is composed of a semantic encoder $E_t(\cdot)$ and a visual encoder  $E_v(\cdot)$ that maps the attribute name and the probe images into a shared semantic-visual space.} \revise{The semantic encoder is a masked self-attention Transformer~\citep{radford2019language} and the visual encoder is a Vision Transformer~\citep{dosovitskiy2020image} ViT-B/32 with 12 layers and the input patch size is 32 $\times$ 32.} 
Then the real-valued confidence $r_a(y)$ is calculated by the similarity measurement $f_{\mathit{sim}}$ as:

\begin{equation}
    r_a(y) = \sum_{i=1}^{m} f_{\mathit{sim}} \bigl( E_t(a), E_v(x_i) \bigl) \,,
    \label{equ:sim}
\end{equation}
where the attribute value $r_a(y)$ indicates the possibility that attribute $a$ would appear in category $y$.
\revise{Following CLIP, we turn a single attribute name~(\eg, ``narrow-road'') into a sentence containing the attribute ~(\eg, ``This photo contains narrow-road'') as the input of $E_t(\cdot)$.} The similarity measurement is the dot product:
\begin{equation}
    f_\mathit{sim} (\alpha,\beta) = <\alpha , \beta> \,.
\end{equation}

In this way, we can easily replace humans in associating each attribute and the corresponding category, and predict the class semantic attributes for each class as $r_\mathcal{A}(y) \in \mathbb{R}^{N_a}$. In spite of decreasing manual labor significantly, the real-valued attribute value works better in discriminating and comparing between different classes than binary attributes annotated by the human. \tworevise{We denote the attributes annotated by CLIP model as Remote Sensing Multi-Modal Attribute~(RSMM-Attribute) in the following text.}

\tworevise{\subsection{Training and fine-tuning of the CLIP model}}
\tworevise{The pre-trained CLIP~\citep{radford2021learning} model is fine-tuned with remote sensing dataset~\citep{lu2017exploring} and the corresponding text descriptions. Here we first introduce the training procedure of the CLIP~\citep{radford2021learning} model as the background for our methodology.} 

\tworevise{The CLIP network should be able to associate visual and semantic information in a common embedding space, for which the training procedure involves pre-training a semantic-visual network with a large-scale dataset containing images and the corresponding text descriptions.} The training procedure is the same as the fine-tune process in Figure~\ref{fig:RSMM}~(left), where a batch of $B$ images $\{x_1, x_2, \dots, x_B\}$ are passed to the visual encoder $E_v(\cdot)$ to extract image representations $\{E_v(x_1), E_v(x_2), \dots, E_v(x_B)\}$ and the text descriptions $\{t_1, t_2, \dots, t_B\}$ are processed by the semantic encoder $E_t(\cdot)$ and output text representations  $\{E_t(t_1), E_t(t_2), \dots, E_t(t_B)\}$. Then a contrastive learning paradigm is adopted where the cosine similarity between positive image-text pairs~(\ie, $x_i$ and $t_i$, where $i \in \{1,2,\dots,B\}$) are optimized to be 1, while the similarity of negative image-text pairs ~(\ie, $x_i$ and $t_j$, where $i \neq j$) are optimized to be close to 0. In particular, the sum of two InfoNCE loss $L_{\scalebox{.6}{NCE}}$ \citep{oord2018representation} is minimized to learn a joint representation of image and texts as follows:
\revise{\begin{align}
\label{equ:NCE}
\begin{split}
    \mathcal{L}_{con} = -\sum_{i=1}^{B}\Bigl(\log{ L_{\scalebox{.6}{NCE}}\bigl(E_v(x_i),E_t(t_j)\bigr)}\\
	+ \log{L_{\scalebox{.6}{NCE}} \bigl( E_t(t_i),E_v(x_j) \bigr) } \Bigr) \,,
\end{split}
\end{align}
where $L_{\scalebox{.6}{NCE}}(E_v(x_i),E_t(t_j))$ denotes the visual to text similarity:}

\revise{\begin{equation}
\label{equ:nce2}
L_{\scalebox{.6}{NCE}}\bigl(E_v(x_i),E_t(t_j) \bigl) = \frac{exp \bigl( E_v(x_i) \cdot E_t(t_i)/ \tau \bigr)}{\sum_{j=1}^{B} exp \bigl( E_v(x_i) \cdot E_t(t_j) \bigr)  } \,,
\end{equation}
and the $L_{\scalebox{.6}{NCE}}\bigl(E_t(t_i), E_v(x_j)\bigl)$ is the text to visual similarity:
\begin{equation}
\label{equ:nce3}
L_{\scalebox{.6}{NCE}} \bigl(E_t(t_i), E_v(x_j)\bigl) = \frac{exp \bigl( E_v(x_i) \cdot E_t(t_i)/ \tau \bigr) }{\sum_{j=1}^{B} exp\bigl(E_t(t_i) \cdot E_v(x_j) \bigr)} \,,
\end{equation}
with $\tau$ being the temperature hyper-parameter.
After pre-training, the text and visual encoder in the CLIP network is able to project the texts and images in one common space.}

Since the pre-training dataset of CLIP~\citep{radford2021learning} mainly contains ordinary optical images collected from the internet, which lacks domain knowledge for remote sensing scenes. In this way, the pre-trained model would suffer from domain gaps. To this end, we fine-tune the CLIP model with remote sensing scene images and corresponding descriptions from RSICD dataset~\citep{lu2017exploring}. The fine-tune procedure is illustrated in Figure~\ref{fig:RSMM}~(left), where a batch of images and corresponding text descriptions are encoded to a common space with the visual encoder $E_v(\cdot)$ and semantic encoder $E_t(\cdot)$. Then we optimize the sum of two InfoNCE loss $L_{\scalebox{.6}{NCE}}$ \citep{oord2018representation} to train the encoders as in Equation~(\ref{equ:NCE}). \tworevise{Afterwards, the CLIP model can be used to map the target image and all attributes into a common visual-semantic space, where the attribute value reflects the strength of association between the attribute and image as in Equation~(\ref{equ:sim}).}



\section{Deep Semantic-Visual Alignment model for Zero-Shot Learning}
\label{sec:DSVA}
We start by formalizing the zero-shot learning~(ZSL) and generalized zero-shot learning~(GZSL) tasks. Then we introduce the architecture of our Deep Semantic-Visual Alignment~(DSVA) model. Afterwards, we introduce the loss functions used to supervise the model training, and describe the inference process for (generalized) zero-shot.

\subsection{Problem definition for (generalized) zero-shot learning}
We are interested in the zero-shot learning problem where the training and test classes are disjoint. The training set is $S=\{x, y, r_\mathcal{A}(y)| x \in \mathcal{X}, y \in \mathcal{Y}^s \}$, which consists of remote sensing image $x$ in the RGB image space $\mathcal{X}$, label $y$ from seen classes $\mathcal{Y}^s$, and \revise{attribute embedding} vector $r_\mathcal{A}(y) \in \mathbb{R}^{N_a}$ derived from Section~\ref{sec:RSMM}. The unseen class label is denoted with $\mathcal{Y}^u$, and the \revise{attribute embedding} vectors for unseen classes $\{r_\mathcal{A}(y)| y \in \mathcal{Y}^u \}$ is also known. The ZSL task aims to predict the label of images from unseen classes, \ie, $\mathcal{X} \rightarrow \mathcal{Y}^u$. While in practice, it is hard to tell if the test image comes from seen or unseen classes. To this end, generalized zero-shot learning~(GZSL) aims to classify images from both the seen and unseen classes, \ie, $\mathcal{X} \rightarrow \mathcal{Y}^u \cup \mathcal{Y}^s$.

\subsection{Deep Semantic-Visual Alignment~(DSVA) model}

\begin{figure*}
    \centering
    \includegraphics[width=.9\linewidth]{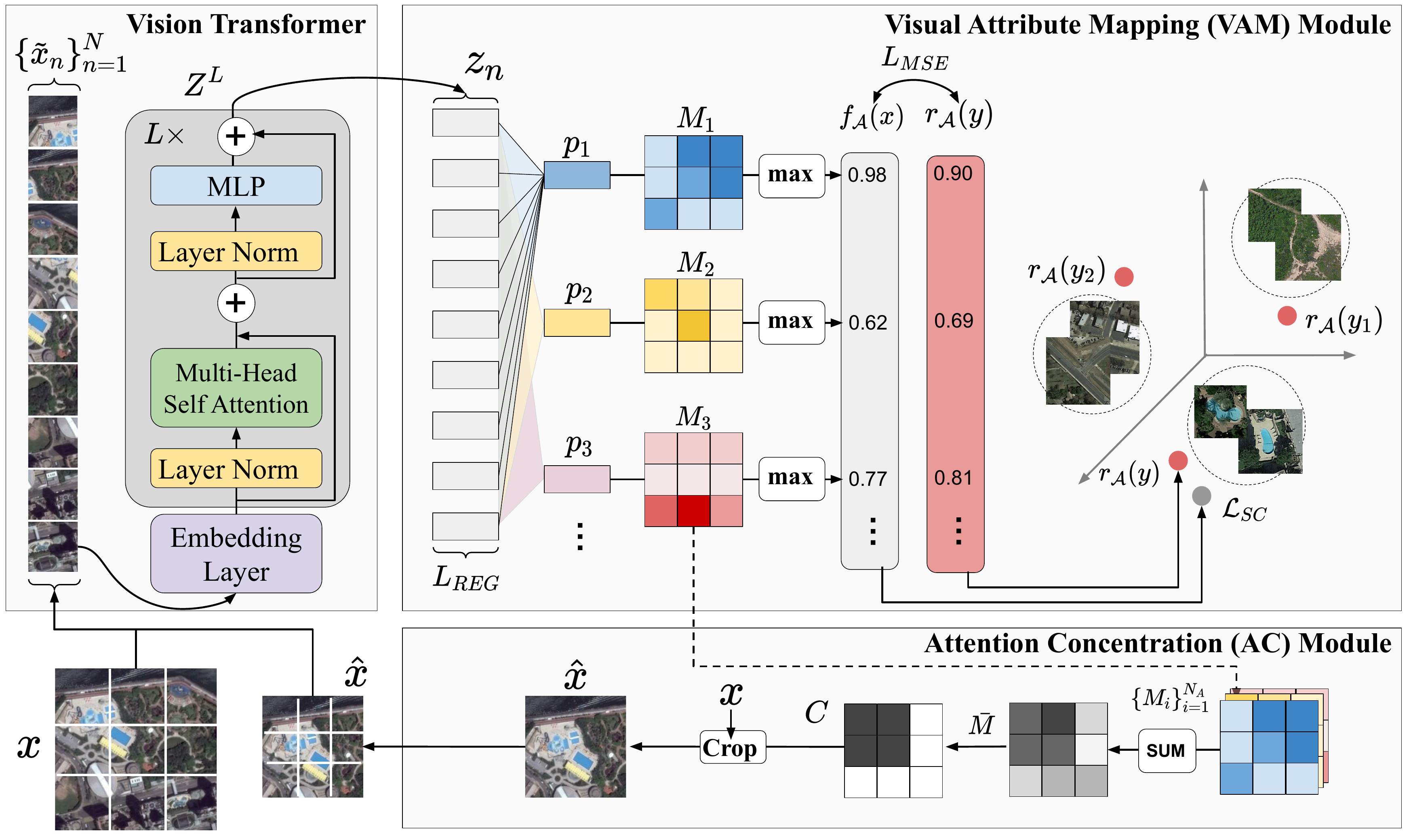}
    \caption{The main architecture of the DSVA model. The proposed DSVA model consists of a \revise{vision transformer} that extracts image features, a Visual Attribute Mapping Module that maps the image into attribute space for ZSL classification, and an Attention Concentration Module that focuses on the informative attribute regions.}
    
    \label{fig:model}
\end{figure*}

As shown in Figure~\ref{fig:model}, the DSVA model utilizes a transformer equipped with self-attention layers to extract the image representations, then maps the representations to the attribute space with a Visual-Attribute Mapping~(VAM) module and finally predicts the class label for each image according to the attribute similarity. Besides, we propose an Attention Concentration~(AC) module to concentrate on the informative image region with the help of attribute-related attention. In the following section, we will formulate the model architecture mathematically.

\subsubsection{Vision transformer} 
Vision transformer use long-range interaction to extract image representations, which is critical for remote sensing scene recognition as the global interaction of different spatial areas in the image contributes to the scene prediction. Different from convolutional neural network that has image-specific inductive bias such as small receptive field, the self-attention mechanism in transformer results in a much larger receptive field and introduces long-range interactions between different image regions. As shown in Figure 5, the input image $x\in \mathbb{R}^{H \times W \times C}$, with $H$, $W$ and $C$ being the height, width and channel, is reshaped to a sequence of image patches $ \{ \tilde{x}_{n} \}_{n=1}^{N}$, \revise{and $\tilde{x}_{n} \in \mathbb{R}^{\frac{H}{k} \times \frac{W}{k} \times C} $.
$k$ is the number of rows (or columns) for patches and $N=k\times k$.} 

\revise{To extract the image representation, we first learn a linear embedding layer $f_0(\cdot)$
to map the image patches to $D$ dimensions patch embeddings $Z^0 \in \mathbb{R}^{N \times D}$:}
\revise{\begin{equation}
Z^0 = [f_0(\tilde{x}_{1}), f_0(\tilde{x}_{2}), \dots, f_0(\tilde{x}_{N})]\,.
\label{equ:self_attention0}
\end{equation}}

Then the patch embeddings will be forward to the transformer encoder consisting of $L$ layers of multi-head self-attention~(MHSA) subnet and multi-layer perceptron~(MLP) subnet, following~\citet{dosovitskiy2020image}:
\begin{align}
Z^{\prime l} = \text{MHSA}\bigl(\text{LN}(Z^{l-1})\bigr) + Z^{l-1}      \\
Z^l = \text{MLP}\bigl(\text{LN}(Z^{\prime l})\bigr) + Z^{\prime l} \qquad 
\end{align}
where $l = 1, \dots, L$ denotes the layers index, and each of the latent embeddings $Z^0, Z^1,\dots, Z^L$ is with shape $\mathbb{R}^{N \times D}$. Where the first input vector $Z^0$ is from Equation~\ref{equ:self_attention0}. Residual connection~\citep{he2016deep} is employed for each of the two subnets followed by layer normalization~(LN) operation~\citep{ba2016layer}.

\revise{In each multi-head self-attention module, with the normalized input vector $\text{LN}(Z^{l-1}) \in \mathbb{R}^{N \times D}$~(we use $Z$ for simplicity), we calculate the weighted sum of each element in the input sequence, where the weight is based on the pairwise similarity between two elements of the sequence. The input sequence $Z \in \mathbb{R}^{N \times D}$ is mapped into three tensors, \ie, the query~(Q), key~(K), and value~(V), by a linear layer, as~\citet{dosovitskiy2020image} do:
\begin{equation}\tag{8}
[Q, K, V] = Z U_{qkv}, \qquad U_{qkv} \in \mathbb{R}^{D \times D_{qkv}}\,.
\label{equ:self_attention0}
\end{equation}}
\revise{Afterwards, we follow \citet{vaswani2017attention} to calculate the self-attention over the query and key via scaled dot product attention:}

\revise{\begin{equation}
\text{Attention}(Q, K) = \text{softmax}(\frac{QK^T}{\sqrt{D_{qkv}} }) \,.
\label{equ:self_attention1}
\end{equation}}

\revise{Then we multiply the attention with the V tensors:
\begin{equation}
\text{SA(Z)} = \text{Attention}(Q, K) V \,,
\label{equ:self_attention2}
\end{equation}
where $\text{SA(Z)}$ is the output of self-attention module.}
\revise{In the multi-head self-attention~(MHSA) subnet, the above self-attention operation is duplicated for $s$ times, namely ``heads'', and the output is concatenated and projected by a linear layer as follows:}

\revise{\begin{equation}
\text{MHSA}(Z) = [\text{SA}_1(Z),  \dots, \text{SA}_s(Z) ] U_{msa},  U_{msa} \in \mathbb{R}^{s \cdot D_{qkv} \times D}\,.
\label{equ:MHSA}
\end{equation}}
\revise{To make sure the number of parameters constant when changing $s$, the $D_{qkv}$ in Equation~(\ref{equ:self_attention0}) and (\ref{equ:MHSA}) is typically set to D/s.
The interaction between all local image patches combines visual cues from two local spots that are far away, thus helps the image representations to incorporate necessary information for image recognition. Given the output image feature $Z^L \in \mathbb{R}^{N \times D}$ with $k \times k$ local embeddings, each local embedding $z_n \in \mathbb{R}^{D}$ encodes the information from local image patch $\tilde{x}_n$ as well as its interaction with all other image patches. }








\subsubsection{Visual-attribute mapping module}

\revise{In this section, we propose a Visual-attribute mapping~(VAM) module to regress the attribute values from the image features, and predict the image category. We construct a visual-attribute mapping layer to learn $N_a$ prototype vector $p_1, p_2, \dots p_{N_a}$. The i-th prototype vector $p_i \in \mathbb{R}^D$ encodes the visual cues for the $i$-th attribute $a$. Note that the prototype vectors are trainable vectors.}

\revise{Since the prototype vector should encode the visual property for each attribute, we use the dot-product as the similarity between the attribute prototype vector $p_i$ and each local image embedding $z_n$, which represents the possibility that image region $\tilde{x}_n$ contain the specific attribute:
\begin{equation}
    \text{sim}(p_i,z_n) = p_i \cdot z_n \,.
\end{equation}}
\revise{As shown in Figure~\ref{fig:model}, we reshape the similarity values between the i-th attribute and $N = k \times k$  local image embeddings to form an attention map $M_i \in \mathbb{R}^{k \times k}$ , indicating the possibility that attribute $a_i$ appearing in the image $x$:}
\begin{equation}
    M_i =  \{p_i \cdot z_n \}_{n=1}^{N} \,.
    \label{equ:M}
\end{equation}
Then we predict the attribute value $f_{a}(x)$ by maximizing the similarity between each image patch and the attribute prototype: 
\begin{equation}
f_{a}(x) = \max_n \{z_n \cdot p_i\} \,,
\end{equation}
where $f_{a}(x) \in \mathbb{R}$ is the predicted attribute value.
Overall, the predicted attribute embedding for image $x$ is $f_\mathcal{A}(x) \in \mathbb{R}^{N_a}$. 
\revise{To assign the image to a specific class, we calculate the compatibility score between the predicted attribute embeddings and the ground truth attribute embedding among all training classes as follows:}
\begin{equation}\tag{15}
    S = f_\mathcal{A}(x) \cdot r_\mathcal{A}(y) \,.
\end{equation}

\subsection{Attention concentrating~(AC) module}
As shown in Figure~\ref{fig:model}, the attention map $M_i$ indicates the image regions that share similar properties with the attribute prototypes, thus taking advantage of these attention maps would help the model locate attribute information and transfer the attribute knowledge between classes. To this end, we propose an attention concentrating module to crop and highlight the attribute informative image regions and train the DSVA network with the cropped images $\hat{x}$ once more. 

Given $N_a$ attribute attention maps $\{M_i\}_{i=1}^{N_a}$ generated from the VAM module~(see equation~\ref{equ:M}), the goal is to concentrate on the attribute related image regions and crop the original image $x$. We first sum the attention maps to get the mean attention:
\begin{equation}
    \bar{M} =  \frac{1}{N_a}\sum_{i=1}^{N_a} M_i \,.
\end{equation}

Then the average attention value is calculated as follows: 
\begin{equation}
    \bar{m} = \frac{1}{N}\sum_{\alpha=1}^{k} \sum_{\beta=1}^{k} {\bar{M}_{\alpha, \beta}} \,,
\end{equation}
with $\alpha$ and $\beta$ being the spatial coordinate of the mean attention, and $N = k \times k$. Then we generate a concentration mask $C$ with size $k \times k$ to highlight the informative image regions,
\begin{equation}
C_{\alpha, \beta} = \left\{\begin{matrix}
 1 & \text{if} \quad \bar{M}_{\alpha, \beta} \geq \bar{m}\\
 0 & \text{if} \quad \bar{M}_{\alpha, \beta} < \bar{m} \,.
\end{matrix}\right.
\end{equation}
Where the regions with attribute attention higher than the average value $\bar{m}$ is marked as one, and the regions with attribute attention value lower than $\bar{m}$ is marked as zero.
Then we use the smallest bounding box that covers all the non-zero values in $C$ to crop the original image $x$ into a cropped image $\hat{x}$, and feed the cropped image into the \revise{vision transformer} and VAM module again. We run the VAM module and the AC module iteratively in each batch, where the AC module will help the network to focus on informative attribute regions that point out the discriminative details between various classes.

\subsection{Training loss}
In this section, we introduce the loss functions $\mathcal{L}_{\Scale[0.6]{\text{DSVA}}}$ to train our DSVA model. 

\subsubsection{Semantic compatibility loss} 
Semantic compatibility loss is used to supervise the training of the DSVA model. Given the input image $x$ with label $y$ and the ground truth \revise{attribute embedding} $r_\mathcal{A}(y)$, we propose to use cross-entropy loss, encouraging the image to have a high compatibility score with its corresponding attribute label as follows:
\begin{equation}
    {L}_{\Scale[0.6]{SC}}(x)  = -\log \frac{\exp\bigl(f_\mathcal{A}(x) \cdot r_\mathcal{A}(y)\bigr)}{\sum_{y_i \in \mathcal{Y}^s}\exp\bigl(f_\mathcal{A}(x) \cdot r_\mathcal{A}(y_i)\bigr)} \,,
        \label{equ:sem}
\end{equation}
where $f_\mathcal{A}(x) \cdot r_\mathcal{A}(y_i)$ is the compatibility score between the target image $x$ and class $y_i$. 
Similarly, the semantic compatibility loss for the cropped image $\hat{x}$ generated from the AC module is
\begin{equation}
    {L}_{\Scale[0.6]{SC}}(\hat{x})  = -\log \frac{\exp\bigl(f_\mathcal{A}(\hat{x}) \cdot r_\mathcal{A}(y)\bigr)}{\sum_{y_i \in \mathcal{Y}^s}\exp\bigl(f_\mathcal{A}(\hat{x}) \cdot r_\mathcal{A}(y_i)\bigr)} \,.
        \label{equ:sem}
\end{equation}
Here we use cross entropy loss to enforce the compatibility score between target image $x$ and its label $y$ to be as high as possible, and the compatibility score between unmatched image and labels to be small.

\subsubsection{Semantic regression loss} 
To facilitate the training of the visual-semantic mapping module, we further consider the attribute prediction as a regression problem and minimize the Mean Square Error~(MSE) between the predicted attribute and the ground truth attribute embedding as follows:
\revise{\begin{equation}
    L_{\Scale[0.6]{MSE}}(x) = \left \| f_\mathcal{A}(x) -r_\mathcal{A}(y) \right \|_2 \,,
\end{equation}}
where $y$ is the label for image $x$. By optimizing the regression loss, we enforce the image representations learned by the transformer to contain semantic information and encode visual cues for each attribute, thus improving the knowledge generalization ability for ZSL.
The semantic regression loss for the cropped image $\hat{x}$ is:
\revise{\begin{equation}
    L_{\Scale[0.6]{MSE}}(\hat{x}) = \left \| f_\mathcal{A}(\hat{x}) -r_\mathcal{A}(y) \right \|_2 \,.
\end{equation}}

Overall, in each batch, our network optimizes the transformer and the visual-attribute mapping module with the following two objective functions iteratively,
\begin{equation}
\begin{aligned}
    \mathcal{L}_{\Scale[0.6]{\text{DSVA}}} = \mathcal{L}_{\Scale[0.6]{SC}}(x) + \lambda \mathcal{L}_{\Scale[0.6]{MSE}}(x) + 
     \mathcal{L}_{\Scale[0.6]{SC}}(\hat{x}) + \lambda \mathcal{L}_{\Scale[0.6]{MSE}}(\hat{x}) \,,
\end{aligned}
\end{equation}
with $\lambda$ being the scaling factors.

\begin{figure*}
    \centering
    \includegraphics[width=.95\linewidth]{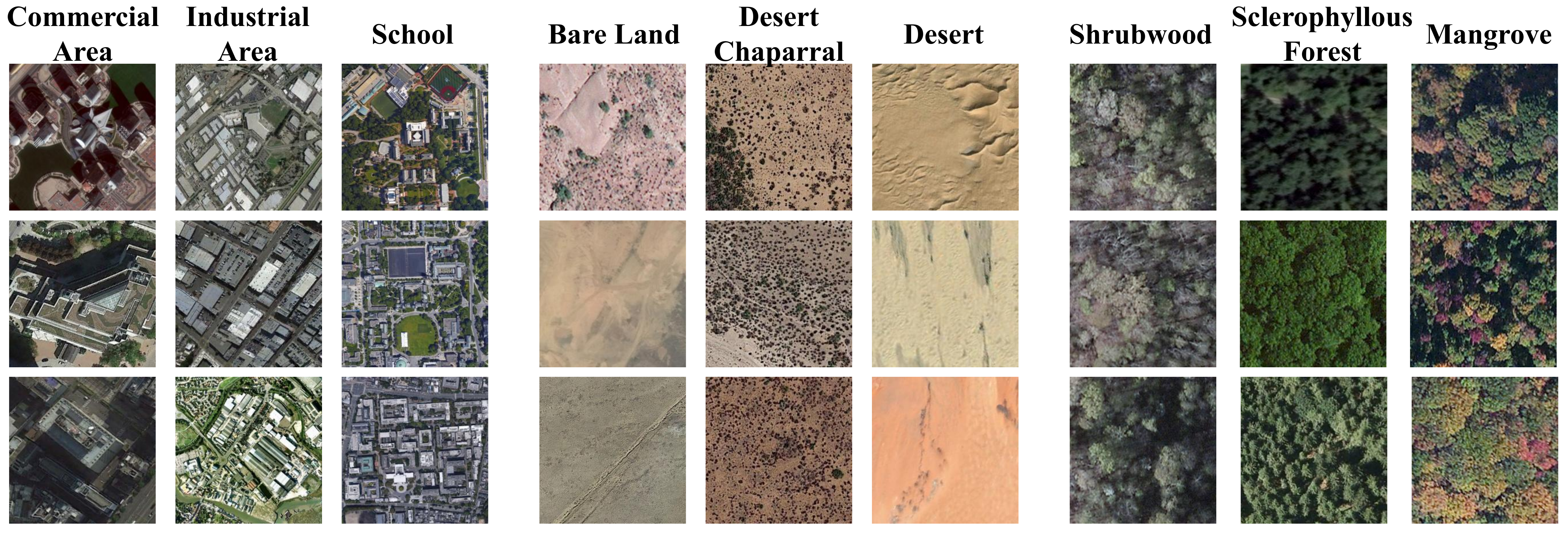}
    \caption{Three groups of similar scene categories from the RSSDIVCS~\citep{li2021learning} dataset. We displace three randomly sampled images from each category.}
    \label{fig:dataset}
\end{figure*}

\subsection{(Generalized) zero-shot inference}
Zero-shot inference classifies the images into unseen classes $\mathcal{Y}^u$. Given input image $x$, the DSVA model first extracts image representations $Z_L$, then maps the visual feature into the attribute space by the VAM module and gets the predicted attribute value $f_{\mathcal{A}}(x)$. In the end, the network searches for the predicted category $\hat{y}$ that has the highest compatibility score with the predicted attribute
\begin{equation}
    \hat{y} = \mathop{\argmax}\limits_{y \in \mathcal{Y}^u} f_\mathcal{A}(x) \cdot r_\mathcal{A}(y) \,.
\end{equation}
For generalized zero-shot inference where the images are classified into both seen and unseen classes, the network searches the predicted category $\hat{y}$ that has the highest compatibility score as follows:
\begin{equation}
    \hat{y} = \mathop{\argmax} \limits_{y \in \mathcal{Y}^u \cup \mathcal{Y}^s} \big( f_\mathcal{A}(x) \cdot r_\mathcal{A}(y)  - \gamma \mathbbm{I}(y \in \mathcal{Y}^s)\big) \,.
\end{equation}
Since the network trained with only seen classes would have bias over training classes, we adopt Calibrated Stacking \citep{chao2016empirical} to decrease the compatibility score of seen classes by a constant indicator $\mathbbm{I}(y \in \mathcal{Y}^s)$, where the indicator is one when $y$ belongs to the seen classes and is zero otherwise, and $\gamma$ is a scaling factor.

\section{Experiment}
\label{sec:Exp}
In this section, we first introduce the dataset and the experiment settings. Then we showcase the ablation study of each component. Afterwards, we compare our DSVA model with other state-of-the-art models on both ZSL and GZSL experiments. Finally, we demonstrate the effectiveness of the RSMM-Attributes with qualitative results.

\subsection{Experiment settings}

\subsubsection{Dataset}
We conduct the zero-shot learning experiment on a widely used large-scale benchmark dataset RSSDIVCS~\citep{li2021learning}, which integrates the image scenes from four datasets, \ie, UCM~\citep{yang2011spatial}, AID~\citep{xia2017aid}, NWPU-RESISC45~\citep{cheng2017remote}, RSI-CB256~\citep{li2017rsi}. The dataset consists of 56,000 images from 70 categories ranging from natural scenes, \eg, lake, mountain, and sea ice, to scenes containing human activity, \eg, school, stadium, and thermal power station. Due to the fine-grained nature of remote sensing scenes where the land covers will appear in different scenes, many categories in the RSSDIVCS dataset are hard to discriminate from each other. We display representative images from three groups of fine-grained scene categories in Figure~\ref{fig:dataset}. We adopt the same dataset split as pioneer works~\citep{li2021robust,li2021representation}, where the 70 categories are randomly split into three different seen/unseen ratios, \ie, 60/10, 50/20, 40/30. Here 60/10 denotes adopting 60 categories as seen classes and the rest 10 categories as unseen classes. 

\revise{We adopt the Remote Sensing Image Captioning Dataset~(RSICD)~\citep{lu2017exploring} to fine-tune the CLIP model. The dataset contains 10,921 remote sensing images collected from Google Earth, Baidu Map, MapABC, and Tianditu. The image size is fixed to 224×224 pixels with various resolutions and there are five text
descriptions for each image. All the images in RSICD dataset are used for the fine-tune procedure.}

\subsubsection{Metrics}
For ZSL task, we adopt the overall accuracy of all unseen classes as the evaluation metric. For the GZSL scenario where the network need to recognize images from both the seen and unseen classes, we follow \citet{xian2018zero} to use the harmonic mean accuracy by considering both the accuracy of seen~(S) and unseen~(U) classes as follows:
\begin{equation}
    H = \frac{(2\times S \times U)}{S + U} .
\end{equation}


\subsubsection{Training details}

\revise{To fine-tune the CLIP model, we adopt the officially pre-trained CLIP~\citep{radford2021learning} model as the backbone. The Adam optimizer with $\beta_1=0.5$ and $\beta_2=0.999$ is adopted to optimize the network. The learning rate is linearly increased from $2 \times 10^{-6}$ to $5 \times 10^{-5}$ for the first 20 epochs to warm up, then decreased by 0.5\% every epoch for 200 epochs. To avoid tuning much hyper-parameters, the temperature hyper-parameter $\tau$ in Equation~(\ref{equ:nce2}) and Equation~(\ref{equ:nce3}), which controls the range of the logits in the softmax function, is directly optimized as a log-parameterized multiplicative scalar following ~\citet{radford2021learning}.
During inference, the number of probe images for each class $m$ in Equation~(\ref{equ:sim}) is set as 10. }

\revise{To train the DSVA network, we adopt the pre-trained Vision Transformer~\citep{dosovitskiy2020image} ViT-B/32 with 12 layers~\citep{radford2021learning} as the backbone. The Adam optimizer with $\beta_1=0.5$ and $\beta_2=0.999$ is adopted to optimize the network in an end-to-end manner. We first fix the transformer backbone, and warm up the Visual-attribute mapping~(VAM) module with 4 epochs by setting the learning rate as $1 \times 10^{-4}$, and then the entire network is trained with a learning rate $10^{-6}$ for 26 epochs.} We set $\lambda=0.08$ for all experiments. The factor for calibrated stacking is set as $1 \times 10^{-4}$.

\subsection{Ablation study}
In this section, we present ablation studies of your proposed DSVA model and the RSMM-Attributes.

\begin{table*}
\caption{\revise{Ablation study over the proposed DSVA model where we report the Top-1~(T1) accuracy on unseen classes for ZSL, as well as the Top-1 accuracy on unseen~(U), seen~(S) and the harmonic mean~(H) for GZSL. We train a single VAM module with only the semantic compatibility loss $\mathcal{L}_{\Scale[0.6]{SC}}$ as the baseline. Note that the last row denotes the full DSVA model, which combines the VAM module and the AC module~(trained with $\mathcal{L}_{\Scale[0.6]{SC}}$, $\mathcal{L}_{\Scale[0.6]{MSE}}$, and $\mathcal{L}_{\Scale[0.6]{AC}}$).}}
\resizebox{\linewidth}{!}{%
\begin{tabular}{l | c | c | c | c  c  c | c  c  c | c   c  c}
\toprule
\multirow{3}{*}{Model} & \multicolumn{3}{c|}{Zero-Shot Learning} & \multicolumn{9}{c}{Generalized Zero-Shot Learning}                                \\ \cline{2-13} 
                       & 60/10       & 50/20       & 40/30      & \multicolumn{3}{c|}{60/10} & \multicolumn{3}{c|}{50/20} & \multicolumn{3}{c}{40/30} \\ 
                       & T1          & T1          & T1         & U       & S      & H      & U       & S      & H      & U       & S      & H      \\ \hline
VAM + $\mathcal{L}_{\Scale[0.6]{SC}}$               &  74.1       & 53.8        &   48.8     & 40.9      & \textbf{79.9}   & 54.1   & 36.1    & \textbf{75.3}   & 48.8   & 31.5    & \textbf{71.2}   & 43.7   \\
$\qquad \quad$ +      $\mathcal{L}_{\Scale[0.6]{MSE}}$             &   80.3       & 63.3        &  56.0       & 62.3    & 72.9   & 67.2   & 50.0      & 62.2   & 55.4   & 42.4    & 58.9   & 49.3   \\
$\qquad \qquad$ +   AC   (DSVA)               &   \textbf{84.0}       & \textbf{64.2}        &    \textbf{60.2}     & \textbf{68.4}    & 67.1   & \textbf{67.7}   & \textbf{53.5}    & 59.8   & \textbf{56.5}   & \textbf{43.7}    & 58.1   & \textbf{49.9 } \\ \bottomrule
\end{tabular}}
\label{tab:ablation_model}
\end{table*}

\subsubsection{Ablation study on the Deep Semantic-Visual Alignment model~(DSVA)} 
To measure the influence of each component of the proposed DSVA model, we design an ablation study. We train the \revise{vision transformer} and the visual-attribute mapping~(VAM) module with only the semantic compatibility loss $\mathcal{L}_{\Scale[0.6]{SC}}$ as the baseline, and train two variants of DSVA by adding the semantic regression loss and the attention concentrate~(AC) module gradually. The model is trained with the RSMM-Attributes automatically annotated by our work.

The ZSL results in Table~\ref{tab:ablation_model}~(left) demonstrate that the full DSVA model improves over the baseline model consistently under different ZSL splits by a large margin. For instance, the accuracy of the baseline model VAM + $\mathcal{L}_{\Scale[0.6]{SC}}$ is improved from 74.1\% to 84.0\%~(60/10), and from 53.8\% to 64.2\%~(50/20), and from 48.8\% to 60.2\%~(40/30). 
In specific, the semantic regression loss $\mathcal{L}_{\Scale[0.6]{MSE}}$ enforces the image representation learned by the transformer backbone to contain semantic information, which boosts the performance by a large margin, \ie,  6.2\%~(60/10), 9.5\%~(50/20), and 7.2\%~(40/30). This indicates that the semantic regression loss encodes the attribute information in image representations and thus improves the knowledge transfer ability of the ZSL model. The attention concentrate module, which helps the network to focus on the informative attribute regions, also provides significant accuracy gain, \ie,  3.7\%~(60/10), 0.9\%~(50/20), and 4.2\%~(40/30). The results indicate that highlighting the attribute related region can help the model to discriminate different classes.

The results under the generalized zero-shot learning~(GZSL) setting are shown in Table~\ref{tab:ablation_model}~(right), which witness a similar trend as the ZSL results. First, introducing the semantic regression loss and the attention concentration module helps the model to recognize unseen classes in the GZSL setting correctly. For instance, the accuracy of unseen classes~(U) is improved from 40.9\% to 68.4\%~(60/10), and from 36.1\% to 53.5\%~(50/20), and from 31.5\% to 43.7\%~(40/30). \revise{Notably, the semantic regression loss improves the performance by a large margin. The reason is that with the semantic regression loss, we enforce the image representations learned by the transformer to contain attribute information and encode visual cues for each attribute prototype, thus facilitating the knowledge transfer between seen and unseen classes.} The results indicate that improving the model's ability to focus on important attributes and informative image regions will significantly decrease the bias on seen classes. Consequently, the seen class accuracy~(S) decreased a bit with our full DSVA model, for which the better model would not have a strong bias on seen classes and result in more balanced accuracy on all classes. \revise{Overall, the harmonic mean~(H) is significantly improved by 13.6\%~(60/10), 7.7\%~(50/20), and 6.2\%~(40/30).}

\begin{figure*}
    \centering
    \includegraphics[width=\linewidth]{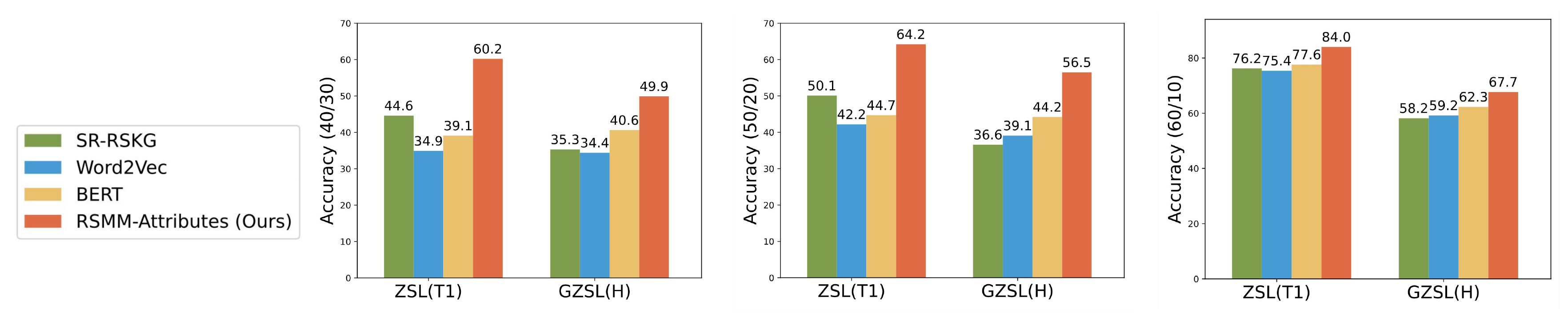}
    \caption{Ablation study over four kinds of class embeddings under three different dataset splits. We report the Top-1~(T1) accuracy of ZSL and the harmonic mean~(H) of GZSL when applying different class embeddings to our proposed DSVA model.}
    \label{fig:ablation_sideinfo}
\end{figure*}

\subsubsection{Ablation study on the class embeddings}
To evaluate the effectiveness of our automatically collected RSMM-Attribute, we compare it with the following three class embeddings widely used by state-of-the-art models. 1) SR-RSKG~\citep{li2021robust} is a semantic representation of RS scenes extracted from knowledge graphs, where 
10 domain experts participated in constructing the knowledge graph. 2) Word2vec embeddings are 300 dimension word embeddings for each RS category, which are extracted from a Word2Vec model pre-trained on Wikipedia corpus. 3) BERT embeddings are built by domain knowledge, where multiple RS experts are invited to depict each RS scene category and summarize them with one sentence. Then the BERT model is utilized to map the sentence to 1024 dimension embeddings.

We train our DSVA network under both ZSL and GZSL settings with the above three class embeddings separately. As shown in Figure~\ref{fig:ablation_sideinfo}, the BERT embeddings are better than the Word2Vec embeddings in all circumstances.
The reason is that the sentences describing each RS category used in BERT embeddings contain more semantic information than only the category name used in Word2Vec embeddings, and results in better generalization ability in the zero-shot learning scenario. 
The SR-RSKG embeddings~\citep{li2021robust} perform similarly to BERT and Word2Vec when trained with our DSVA models. Moreover, our RSMM-Attributes work much better than other alternative class embeddings under both ZSL and GZSL settings with different dataset splits. For instance, with a split 40/30, the model trained with RSMM-Attributes provides significant performance gain compared with the model trained with BERT by 21.1\%~(ZSL) and 9.3\%~(GZSL). 
With a split 50/20, RSMM-Attributes achieve 64.2\%~(ZSL)) and 56.5\%~(GZSL), while BERT only gets 44.7\%~(ZSL) and 44.2\%~(GZSL). 
\revise{The results demonstrate two advantages of the RSMM-Attributes. Firstly, the visual properties encoded in the RSMM-Attributes help the ZSL network to model the visual space of unseen classes and recognize unseen images accurately. Secondly, different from other class embeddings that cannot tell the concrete semantics they encode, each dimension in our RSMM-Attributes denotes a specific semantic attribute, which is intuitive for the ZSL network to link different categories with those attributes and benefit the intra-class knowledge transfer. The results verifies the importance of using language-visual multi-modal network for annotating RSMM-Attributes. }

\subsubsection{Ablation study on the attention concentration module}
\begin{table}[tb]
\centering
\caption{\revise{Comparison between our attention concentration module with other attention modules, where we report the Top-1~(T1) accuracy on unseen classes for ZSL. We train the DSVA model with the semantic compatibility loss $\mathcal{L}_{\Scale[0.6]{SC}}$ as the basemodel. Then we compare four different attention modules.} }
\resizebox{\linewidth}{!}{%
\begin{tabular}{l | c | c | c }
\toprule
\multirow{3}{*}{Model} & \multicolumn{3}{c}{Zero-Shot Learning}                              \\ \cline{2-4} 
                       & 60/10       & 50/20       & 40/30       \\ 
                       & T1          & T1          & T1          \\ \hline
basemodel           &   74.1       & 53.8        &  48.8      \\
basemodel + BAM~\citep{park2018bam}        &   75.5       &   54.6      &  49.0     \\
basemodel + CBAM~\citep{woo2018cbam}        &   75.9       &   54.9      &  49.5    \\
basemodel + GradCAM~\citep{rs12091366}        &   76.8      &  55.2      &  50.1    \\ 
basemodel + Ours       &   \textbf{84.0}       & \textbf{64.2}        &    \textbf{60.2}   \\ \bottomrule
\end{tabular}}
\label{tab:attention}
\end{table}

\revise{We compare our attention concentration module with three other attention modules proposed recently. Bottleneck Attention Module~(BAM)~\citep{park2018bam} infers an attention map along two separate pathways, \ie the channel and spatial attention, to concentrate on the important image regions and channels. Convolutional Block Attention Module~(CBAM)~\citep{woo2018cbam} further utilize maxpooling to generate more salient features from the feature map, to concentrate on the global regions. ~\citet{rs12091366} propose to use Gradient Attention Map~(GradCAM) as the attention module to pay attention to the salient image regions. We train the DSVA model with the semantic compatibility loss $\mathcal{L}_{\Scale[0.6]{SC}}$ as the basemodel, then add the three attention modules separately to verify their effectiveness. As shown in Table~\ref{tab:attention}, our attention concentration module with the semantic regression loss $\mathcal{L}_{\Scale[0.6]{MSE}}$ outperforms other competitors by a large margin. For instance, compared to the GradCAM attention module, our attribute attention concentration module gains 7.2\%~(60/10), 9.0\%~(60/10), and 10.1\%~(40/30). The reason is that our model encodes the attribute information in the image representations and thus improves the knowledge transfer ability of the ZSL model. Besides, the attention concentrate module helps the network to focus on all the informative attribute regions guided by the attribute value for each class, which provides significant accuracy gain compared to other attention modules that only concentrate on one image region inferred by a learnable parameter.}

\begin{table*}[tb]
\centering
\caption{\tworevise{We compare the GZSL performance of our DSVA model with five SOTA models, where the models are trained with three different kinds of class embeddings~(SideInfo), \ie, SR-RSKG~\citep{li2021robust}, Word2Vec~\citep{w2v}, and BERT~\citep{devlin2018bert}. The results are harmonic mean~(H). We build DSVA model with two different backbones, \ie, ResNet18~\citep{he2016deep}~(denoted as RN18) and  Vision Transformer~(denoted as ViT)~\citep{dosovitskiy2020image}.}}
\resizebox{\linewidth}{!}{%
\begin{tabular}{l | c |  c |  c |  c |  c |  c |  c  |  c }
\toprule
SideInfo                  & Seen/Unseen ratio & SAE & DMaP & CIZSL & CADA-VAE & DAN   & DSVA~(RN18)~(Ours) & DSVA~(ViT)~(Ours)  \\ \hline 
\multirow{3}{*}{SR-RSKG}  & 60/10             & 28.9 & 30.1 & 23.7  & 38.1     & 40.3 &  50.7 &     \textbf{58.2}     \\
                          & 50/20             & 23.7 & 23.4 & 13.9  & 32.9     & 34.1 &  36.0  &       \textbf{36.6}    \\
                          & 40/30             & 16.9 & 16.2 & 8.1   & 28.1     & 29.6 & 33.9  &    \textbf{35.3} \\ \hline
\multirow{3}{*}{Word2Vec} & 60/10             & 28.0   & 28.9 & 25.2  & 32.9     & 34.1 &    48.2    &     \textbf{59.2}    \\
                          & 50/20             & 21.0   & 20.3 & 15.7  & 30.3     & 31.4 &   36.4    &      \textbf{39.1}   \\
                          & 40/30             & 17.2 & 16.8 & 9.1   & 26.1     & 25.6 &   30.1   &      \textbf{34.4}     \\\hline
\multirow{3}{*}{BERT}     & 60/10             & 28.6 & 26.6 & 25.0    & 36.3     & 38.0   &   52.0   &      \textbf{62.3}     \\
                          & 50/20             & 21.5 & 19.5 & 15.0    & 31.5     & 31.5 &   39.8    &      \textbf{44.2}    \\
                          & 40/30             & 16.7 & 16.3 & 8.6   & 27.1     & 28.2 &   33.8    &      \textbf{40.6}    \\ \bottomrule       
\end{tabular}}
\label{tab:main_sideinfo_gzsl}
\end{table*}

\begin{table*}[tb]
\centering
\caption{\tworevise{We compare the ZSL performance of our DSVA model with seven SOTA models, where the models are trained with three kinds of class embeddings~(SideInfo), \ie, SR-RSKG~\citep{li2021robust}, Word2Vec~\citep{w2v}, and BERT~\citep{devlin2018bert}.  We report the Top-1 accuracy of all unseen classes. We build DSVA model with two different backbones, \ie, ResNet18~\citep{he2016deep}~(denoted as RN18) and ViT~\citep{dosovitskiy2020image}.}}
\resizebox{\linewidth}{!}{%
\begin{tabular}{l | c |  c |  c |  c |  c |  c |  c |  c |  c  |  c }
\toprule
SideInfo                  & Seen/Unseen ratio & SAE  & DMaP & SPLE & CIZSL & CADA-VAE & ZSC-SA & DAN  & DSVA~(RN18)~(Ours) & DSVA~(ViT)~(Ours) \\ \hline
\multirow{3}{*}{SR-RSKG }  & 60/10                & 22.1  & 33.1  &  28.5 & 18.2  & 50.5  & 31.3  &  53.3 &      58.7 & \textbf{76.2}    \\
                          & 50/20            &  12.8 &  20.3 & 17.2  &  8.9 &  39.6 &  19.1 &  45.2 &     46.6 & \textbf{50.1}     \\
                          & 40/30               &  9.2 &  12.9 &  10.2 & 7.1  &  28.2 &  13.6 & 33.4 &    35.9 & \textbf{44.6}      \\ \hline
\multirow{3}{*}{Word2Vec} & 60/10             & 23.5   & 26.0 & 20.1  & 20.6     & 41.4 &  26.7 & 44.3  &  55.7  &   \textbf{75.4}       \\
                          & 50/20             &  13.7 &  16.7 &  13.2 &  10.6 &  30.3 &  15.2 &  34.7 &   39.9   &    \textbf{42.2}    \\
                          & 40/30           & 9.6  & 10.4  & 9.8  & 6.0  & 21.2 &  12.1 &  24.3 &   31.0   &    \textbf{34.9}      \\ \hline
\multirow{3}{*}{BERT}     & 60/10             & 22.0  & 16.4  &  19.0 &  20.4 & 48.1  & 29.3  &  50.2 &  59.4    &   \textbf{77.6}      \\
                          & 50/20               & 12.4  & 15.6  &  13.2 &  10.3 & 37.1  &  18.3 &  43.4 &   44.0    &  \textbf{44.7}     \\
                          & 40/30             &  8.8 & 10.0  & 8.3  &  6.2 &  26.3 & 13.1  & 31.5  &   35.3     &  \textbf{39.1}     \\ 
 \bottomrule       
\end{tabular}}
\label{tab:main_sideinfo_zsl}
\end{table*}

\subsection{Main results}

In this section, we compare our Deep Semantic-Visual Alignment~(DSVA) model with two groups of state-of-the-art models. Non-generative models learns projection function between the image features and class embeddings, \ie, SPLE~\citep{tao2017semantics}, DMaP~\citep{DMAP}, and LPDCMENs~\citep{li2021learning}. Generative models learns auto-encoder or generative adversarial network to synthesize image features for unseen classes, \ie,  SAE~\citep{kodirov2017semantic}, ZSC-SA~\citep{quan2018structural}, CIZSL~\citep{elhoseiny2019creativity}, CADA-VAE~\citep{schonfeld2019generalized}, TF-VAEGAN~\citep{narayan2020latent},  CE-GZSL~\citep{han2021contrastive}, and DAN~\citep{li2021robust}. \revise{We build our DSVA model with two different backbones, \ie, ResNet18~\citep{he2016deep} and ViT~\citep{dosovitskiy2020image}, and all other ZSL models use image representations extracted from ResNet18.} \tworevise{We firstly compare those models trained with three different class embeddings, \ie, SR-RSKG~\citep{li2021robust}, Word2Vec~\citep{w2v}, and BERT~\citep{devlin2018bert}, then we compare the performance of our DSVA model with the best performance of other SOTA models.}

Table~\ref{tab:main_sideinfo_gzsl} displays the generalized zero-shot learning performance of different models trained with three class embeddings. As can be observed, our DSVA model yields a better harmonic mean than all other SOTA models. \revise{Specifically, when trained with BERT embeddings, our DSVA model using ResNet 18 as backbone
achieves 52.0\%~(60/10), 39.8\%~(50/20), and 33.8\%~(40/30), 
which is much higher than the second best model DAN~\citep{li2021robust}, which obtains 38.0\%~(60/10), 31.5\%~(50/20), and 28.2\%~(40/30).} 
\revise{When trained with SR-RSKG embeddings learned from knowledge graphs, our DSVA~(RN18) model still leads to significant performance gain compared to DAN, by 10.4\%~(60/10), 1.9\%~(50/20), and 4.3\%~(40/10). When using a \revise{vision transformer}~(ViT) as the backbone, our performance is further boosted.} This indicates that our DSVA model that enforces the alignment between visual features and class embeddings is able to balance the performance of unseen and seen classes, and decrease the bias towards seen classes. 

\begin{table*}[tb]
\centering
\caption{\tworevise{\revise{Comparing our DSVA model with SOTA models. We build DSVA model with two different backbones, \ie, ResNet18~\citep{he2016deep}~(denoted as RN18) and ViT~\citep{dosovitskiy2020image}. We report the Top-1~(T1) accuracy of unseen classes under the ZSL setting and the harmonic mean~(H) of both seen and unseen classes under the GZSL setting.} \tworevise{ For fair comparison, all the models are trained with SR-RSKG~\citep{li2021robust} as the class embedding~(SideInfo). To verify the effectiveness of our RSMM-Attributes, we further report the results of our DSVA model trained with RSMM-Attributes.} \revise{Some results are``-'' since we cannot access to their official code.}}}
\resizebox{\linewidth}{!}{%
\begin{tabular}{c| c | c | c | c | c | c | c | c }
\toprule
\multirow{3}{*}{SideInfo} & \multirow{3}{*}{Model} & \multicolumn{3}{c|}{ZSL Accuracy} & \multicolumn{3}{c|}{GZSL Accuracy} & \multirow{3}{*}{Model Size}   \\ \cline{3-8} & 
                       & 60/10       & 50/20       & 40/30     & 60/10       & 50/20       & 40/30   & \\ 
                      &  & T1          & T1          & T1        & H       & H     & H     &  (MB)\\ \hline
\multirow{10}{*}{SR-RSKG} & SAE~\citep{kodirov2017semantic}&  22.1 &  12.8 &  9.2 & 28.9  & 23.7  &  16.9  & 44.59\\ \cline{2-9}
& CIZSL~\citep{elhoseiny2019creativity}& 18.2  & 8.9  &  7.1 & 23.7 & 13.9  &  8.1  &  50.59 \\ \cline{2-9}
 & DMaP~\citep{DMAP}& 33.1  & 20.3  &  12.9 & 30.1  & 23.4  &  16.2  & 44.59 \\ \cline{2-9}
 & CADA-VAE~\citep{schonfeld2019generalized}&  50.5 & 39.6  &  28.2 &  38.1 & 32.9  &  28.1  & 26.34 \\ \cline{2-9}
& TF-VAEGAN~\citep{li2021robust}&  51.5 & 41.9  & 30.0  &  40.1 & 35.0  & 29.2  & 291.73 \\ \cline{2-9}
& CE-GZSL~\citep{li2021robust}&  53.6 & 44.7  & 32.1  &  42.9 & 35.9  & 32.1 & 156.08 \\ \cline{2-9}
 & ZSC-SA~\citep{quan2018structural}&  31.3 & 19.1  &  13.6 & -  & -  &  -  & -\\ \cline{2-9}
 & LPDCMENs~\citep{li2021learning}&  43.8 & 24.9  & 21.6  &  - & -  & -   & -\\ \cline{2-9}
  & DAN~\citep{li2021robust}&  53.3 & 45.2  & 33.4  &  40.3 & 34.1  & 29.6   & -\\ \cline{2-9}
& DSVA~(RN18)~(Ours) & 58.7  &  46.6 &  35.9 &  50.7 &  36.0 & 33.9   & 43.01 \\ \cline{2-9}
 & DSVA~(ViT)~(Ours) & 76.2  &  50.1 &  44.6 &  58.2 &  36.6 & 35.3   & 334.20 \\ \hline
RSMM- & DSVA~(RN18)~(Ours) & 69.8  &  48.7 &  36.4 &  52.1 &  37.6 & 37.3   & 43.01 \\ \cline{2-9}
Attributes & DSVA~(ViT)~(Ours) &  \textbf{84.0} & \textbf{64.2}  &  \textbf{60.2} & \textbf{67.7}  & \textbf{56.5}  & \textbf{49.9}   & 334.20 \\
\bottomrule
\end{tabular}}
\label{tab:main_result}
\end{table*}

Table~\ref{tab:main_sideinfo_zsl} displays the Top-1 ZSL accuracy of different models trained with two class embeddings, where our performance is comparable to or better than other SOTA methods. Notably, when trained with Word2Vec and BERT embeddings, 
our DSVA model outperforms other models by a large margin, \eg, up to 30\%. 
\tworevise{For instance, the DSVA~(RN18) model trained with Word2Vec improves the accuracy of DAN~\citep{li2021robust} from 44.3\%~(60/10) to 55.7\%~(60/10). 
The DSVA~(RN18) model trained with BERT embeddings improves the accuracy of CADA-VAE~\citep{schonfeld2019generalized} from 48.1\%~(60/10) to 59.4\%~(60/10).}
The impressive improvement demonstrates the ability of our model to recognize unseen classes with the help of different class embeddings. 
\begin{figure*}
    \centering
    \includegraphics[width=\linewidth]{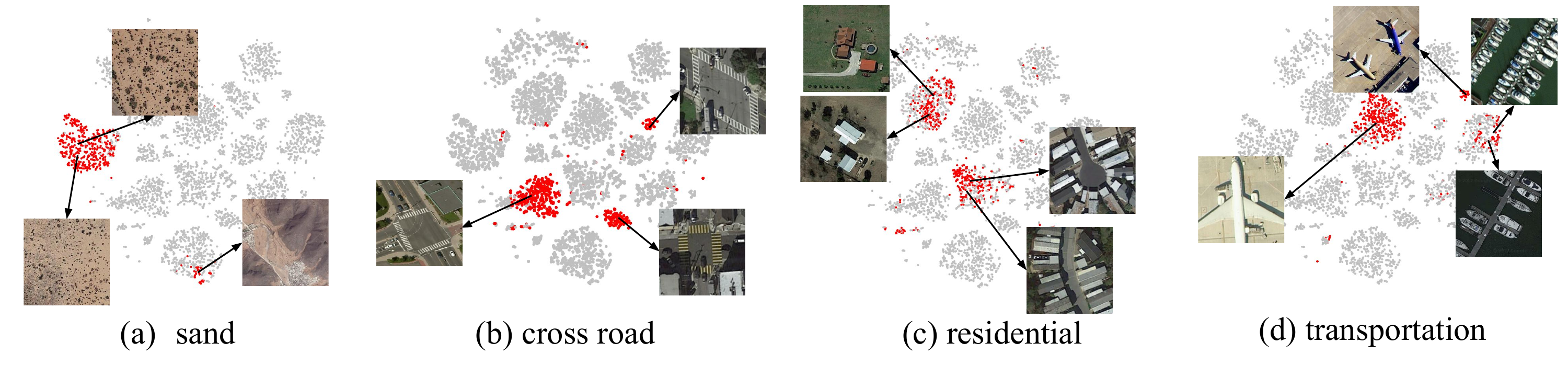}
    \caption{The t-SNE visualization for four attributes. The red dots indicate images that activate the attribute, and we show several image examples in the cluster center.}
    \label{fig:attribute_tsne_2}
\end{figure*}

\tworevise{In Table~\ref{tab:main_result}, we compare our DSVA model trained with RSMM-Attribute and the best performance of other SOTA models. For fair comparison, we train our model with two different class embeddings, \ie, SR-RSKG~\citep{li2021robust} and our RSMM-Attributes. Besides, We build our DSVA model with two different backbones, \ie, ResNet18~\citep{he2016deep}~(denoted as RN18) and  Vision Transformer~(denoted as ViT)~\citep{dosovitskiy2020image}. Under the ZSL setting, compared with all other state-of-the-art models, our model yields consistent improvement on three dataset splits. Compared to the recent proposed non-generative model LPDCMENs designed for ZSL remote sensing scene classification, our DSVA~(RN18) trained with SR-RSKG gained 14.9\%~(60/10), 21.7\%~(50/20), and 14.3\%~(40/30), respectively. When trained with our RSMM-Attributes, the ZSL performance of our DSVA model is further improved. Our DSVA~(RN18) improves generative model DAN that synthesizes images for unseen classes from 53.3\% to 69.8~(60/10), from 45.2\% to 48.7\%~(50/20), and from 33.4 to 40.5~(40/30)\%.} Compared with the SOTA ZSL model CE-GZSL~\citep{han2021contrastive} designed for ordinary optical images, our model DSVA trained with a light backbone ResNet18 already significantly improves the accuracy with only 43.01 MB parameters, \ie, our model achieves 69.8\%~(60/10), 48.7\%~(60/10), and 36.4\%~(40/30) for ZSL accuracy, while the CE-GZSL model only achieves 53.6\%~(60/10), 44.7\%~(60/10), and 32.1\%~(40/30) with much more parameters~(156.08 MB). When applying a larger model ViT with 334.20 MB parameters as the backbone, the ZSL accuracy of our model is significantly boosted to 84.0\%~(60/10), 64.2\%~(60/10), and 60.2\%~(40/30). The results indicate that our model can already outperform all other ZSL models with very few parameters.
When using a \revise{vision transformer}~(ViT) as the backbone, associating the global information of remote scene images is quite useful for zero-shot model to recognize unseen classes. The same trend is observed under the GZSL setting in Table~\ref{tab:main_result}~(right), where the DSVA~(ViT) model yields the best performance over all other SOTA models. It indicates that even under the realistic setting where the model needs to recognize seen and unseen classes simultaneously, our non-generative model DSVA still yields the best generalization ability and outperforms other generative models.


\begin{figure}
    \centering
    \includegraphics[width=.95\linewidth]{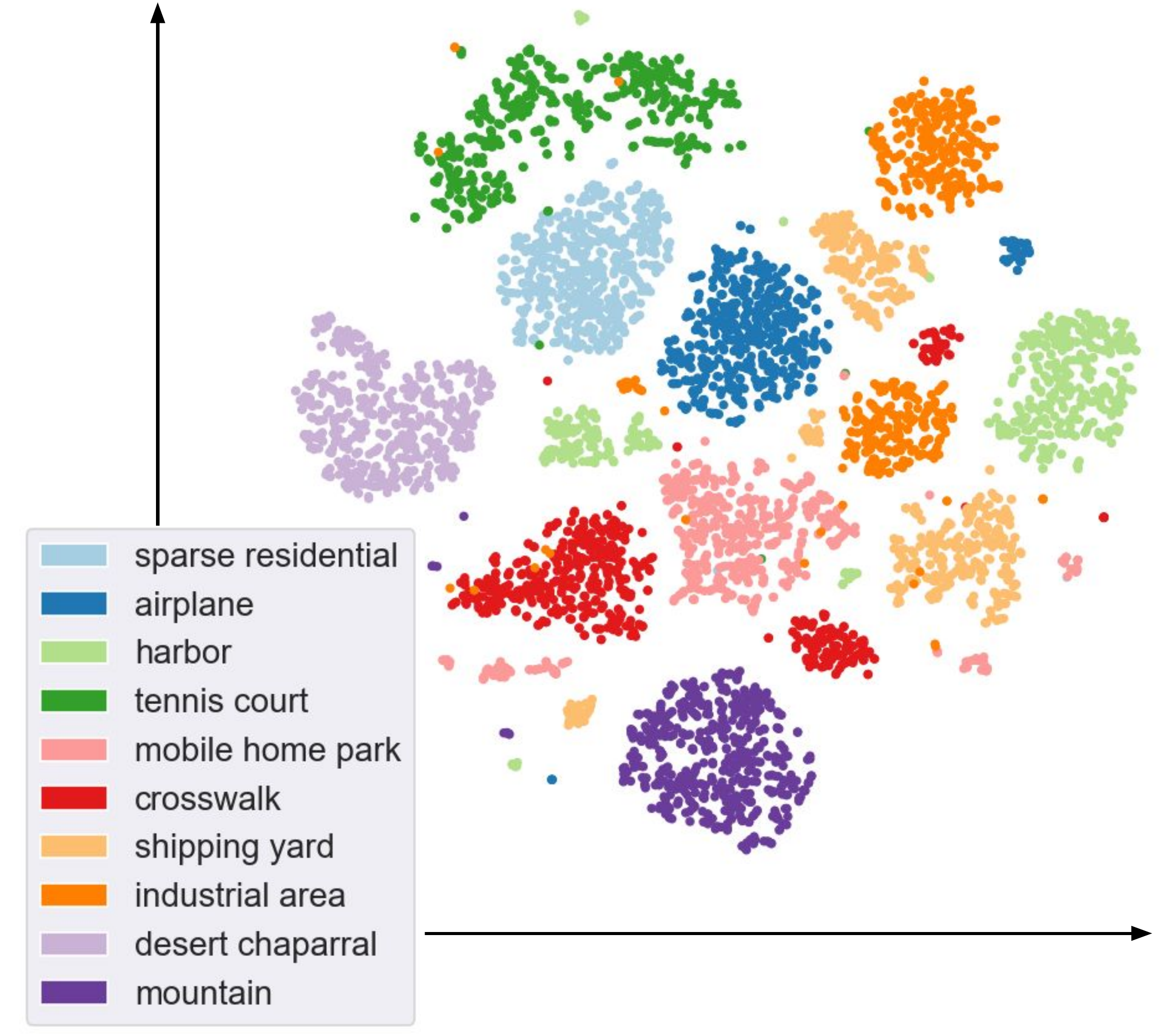}
    \caption{Exploring scene images in 2-D attribute space, where dots with different colors represent images from various categories, and the visualization is finished by t-SNE~\citep{van2008visualizing}.}
    \label{fig:class_tsne}
\end{figure}

\begin{figure}
    \centering
    \includegraphics[width=\linewidth]{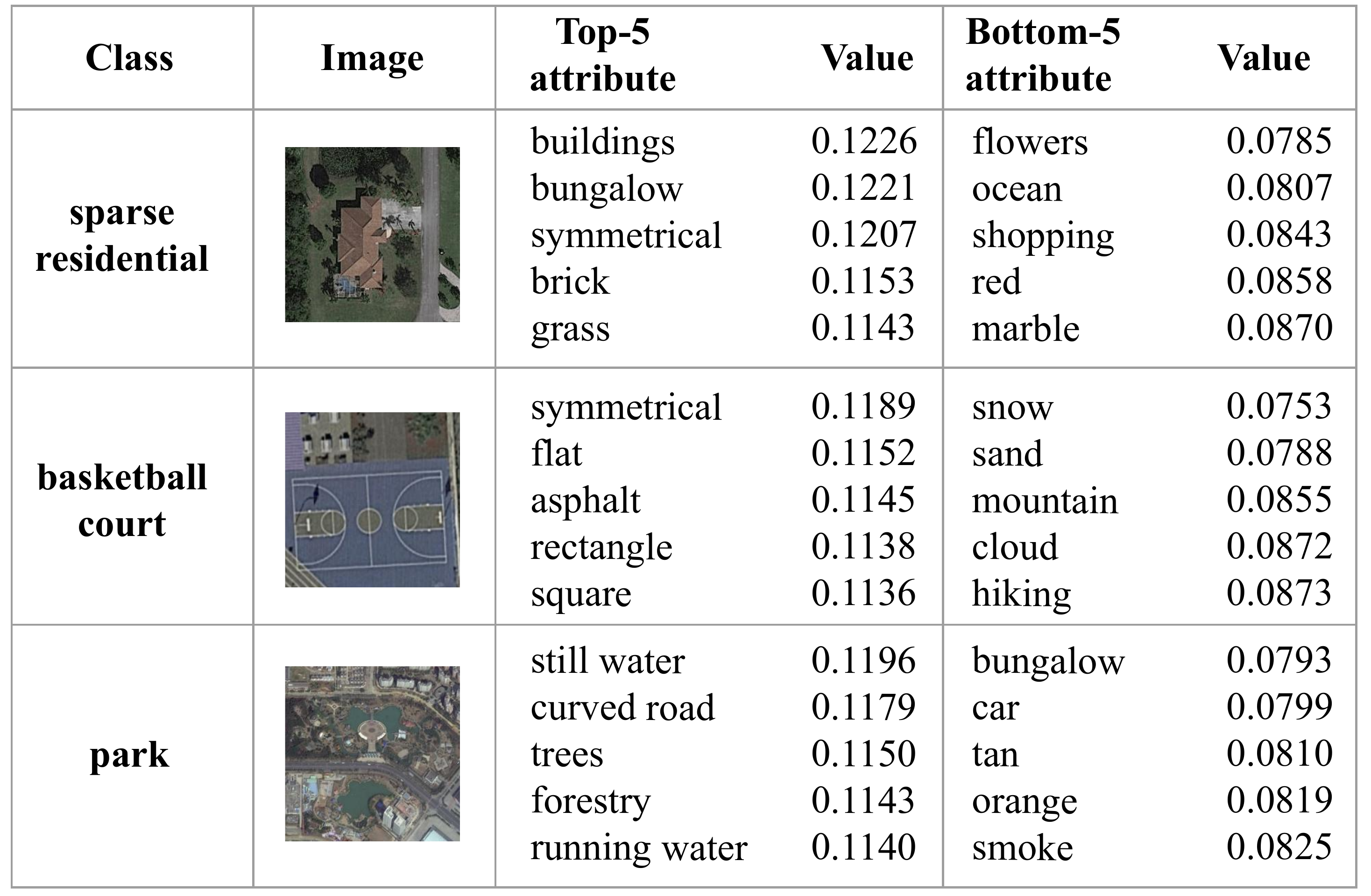}
    \caption{\tworevise{Examples of the ground truth RSSM-Attributes embeddings. We display the example image, attribute name, and attribute value for three classes. Top-5 attributes denote the attributes with the highest attribute value for a certain class, and Bottom-5 attributes denote those with the lowest attribute value.}}
    \label{fig:attribute_vis}
\end{figure}

\begin{figure*}
    \centering
    \includegraphics[width=\linewidth]{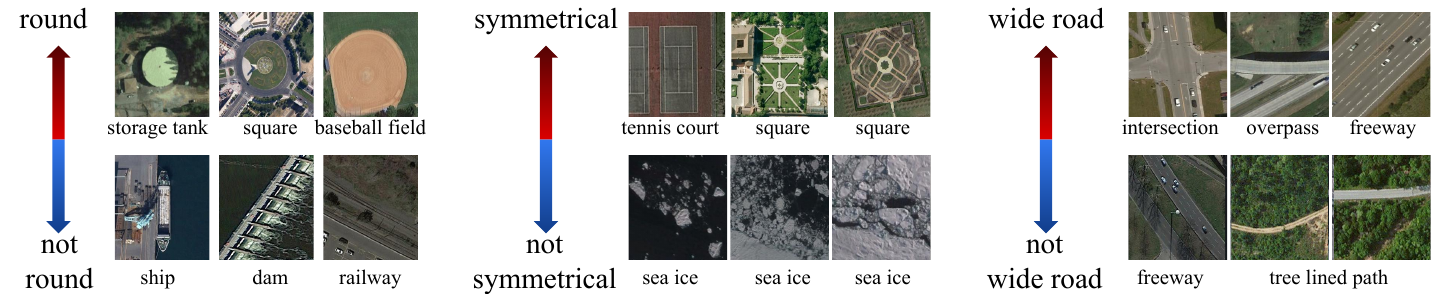}
    \caption{Example (de-)activated images for three RSMM-Attributes, \ie, ``round'', ``symmetrical'', and ``wide road''. For each attribute, the images near the red arrow indicate the positive samples that have the highest attribute value, while images near the blue arrow are negative samples having the lowest attribute value. The text under each image denotes the category label for them.}
    \label{fig:image_attri}
\end{figure*}

\section{Attribute Visualization}
\label{sec:vis}
We start by visualizing the class distribution in the attribute space in Figure~\ref{fig:class_tsne}, where 10 unseen classes are selected with 1,000 images per class. For each image $x$, we extract the RSMM-Attribute value $r_{\mathcal{A}}(x) \in \mathbb{R}^{N_\mathcal{A}}$, where each dimension indicates the similarity $f_{\mathit{sim}}\bigl( E_t(a), E_v(x) \bigl) $ between the image $x$ and the corresponding attribute $a$. Then t-SNE~\citep{van2008visualizing} is used to embed the attribute for all 10,000 images into a 2-D space. Figure~\ref{fig:class_tsne} demonstrates that the attribute space is class discriminative, and various categories can be well separated apart. Besides, attributes can reflect both visual and semantic relatedness. For instance, the visually similar classes, \eg, \textit{mobile home park} and \textit{crosswalk}, locate near each other in the attribute space. 
This is because the RSSM-Attributes can encode not only the visual information in each image, but also the semantic relatedness according to the text description simultaneously. This property can benefit the zero-shot generalization, which transfers visual knowledge with semantic class embeddings.

Then we explore where images with different attributes live in the attribute space. In Figure~\ref{fig:attribute_tsne_2}, we use red dots to represent images that activate a certain attribute, \ie, the top 10\% images that have the highest attribute value, and mark other images with gray color dots. The visualization demonstrated that images containing the same attributes tend to live together in the 2-D attribute space. Meanwhile, images sharing some same attributes may differ from each other according to other attributes. As shown in the figure, the image clusters for each attribute are usually split apart according to their overall appearance similarity. For instance, in Figure~\ref{fig:attribute_tsne_2}~(a), both dessert and mountain contain attribute ``sand'', and in Figure~\ref{fig:attribute_tsne_2}~(d), both airport and  harbor has the attribute ``transportation''. This indicates that the RSMM-Attribute can not only link images sharing the same attribute together, but also discriminate images from various categories apart.

\tworevise{We display some examples of the ground truth RSSM-Attributes embedding for several classes in Figure~\ref{fig:attribute_vis}. Since there are 98 attributes for each class, to save space, we display the top 5 attributes with the highest attribute value and the bottom 5 attributes with the lowest attribute value. As shown in Figure~\ref{fig:attribute_vis}, the top 5 attributes indicate the most representative visual and semantic properties for each class. For example, for the class basketball court, attributes ``symmetrical'' and ``flat'' represents the texture,  ``asphalt'' represents the materials, ``rectangle'' and ``square'' represents the shape. Conversely, attributes that are not commonly associated with basketball courts, such as ``snow'', ``sand'', ``mountain'', ``cloud'', and ``hiking'' have low attribute values.}

In Figure~\ref{fig:image_attri}, we display some image examples of our annotated RSMM-Attributes. For each attribute, we select images that are annotated with the highest attribute value, and images with the lowest attribute value. We can observe the following. First, the activated images that have the highest attribute value all convey the attribute properties correctly, \eg, the ``wide road'' existing in \textit{intersection}, \textit{overpass}, and \textit{freeway}, and the ``symmetrical'' shapes in the \textit{tennis court} and \textit{square}. Moreover, instead of activating images with the same visual appearance, the positive images contain attributes with various visual cues. For instance, though indicating different objects, the round roof in the \textit{storage tank}, the circle road in the \textit{square}, and the round playing ground in the \textit{baseball field} all activate the ``round'' attributes. This observation demonstrates that our model can discover the same attribute on different objects and facilitate attribute sharing and knowledge transfer across classes. In addition, the negative examples for each attribute are not conveying arbitrary properties but show semantically opposite properties. For example, the negative examples for ``round'' are images containing straight stripes, and the negative examples for ``wide road'' are scenes holding the narrow road. This interesting observation indicates the advantage of using the semantic-visual pre-training network, which encodes fluent semantic information visual attribute space and is beneficial for zero-shot learning where each category is described by semantic attributes.

\section{Conclusion}
\label{sec:Conclu}
Driven by the increasing demand for recognizing previous unseen classes in RS scenarios, we aim to improve the performance of ZSL models for RS scene classification in this work. To alleviate the manual effort needed in attribute annotation, we propose a semantic-visual multi-modal network to annotate visually detectable attributes for each category automatically. Moreover, a Deep Semantic-Visual Alignment model is proposed to map visual images into the attribute space and classify images from both seen and unseen classes simultaneously. We explicitly encourage the model to associate local image regions together for better representation learning with the help of self-attention. Moreover, an attention concentration module is proposed to focus on the informative attribute regions. With extensive experiments, we demonstrate that our model improves over the state-of-the-art model by a large margin. Moreover, we qualitatively verify that the RSMM-Attributes annotated by our network are both class discriminative and semantic related, which benefits the zero-shot knowledge transfer between seen and unseen classes.

\section*{Acknowledgement}
This paper was supported in part by the National Key Research and Development Program of China under Grant 2021YFB2900200, National Natural Science Foundation of China under No. 61925101, and 61831002, 61921003, the Beijing Municipal Science and Technology Project NO. Z211100004421017.

\bibliographystyle{cas-model2-names}

\bibliography{mybib}

\end{sloppypar}
\end{document}